\documentclass[conference]{IEEEtran}
\IEEEoverridecommandlockouts
\usepackage{cite}
\usepackage{epsfig,rotating,setspace,latexsym,amsmath,epsf,amssymb,bm,amsfonts}
\usepackage{amsthm}
\usepackage{cite,graphicx,color}
\usepackage{algorithmic}
\usepackage{caption}
\usepackage{verbatim}
\usepackage{mathtools}
\usepackage{calc}
\usepackage{array,multirow}
\usepackage{balance}

\DeclareMathOperator*{\argmax}{arg\,max}

\DeclareMathAlphabet{\pazocal}{OMS}{zplm}{m}{n}

\usepackage{algorithm}
\usepackage{booktabs}

\usepackage{multicol}
\usepackage{tabularx}
\usepackage[table]{xcolor}
\usepackage{subcaption}
\usepackage{textcomp}
\usepackage{xcolor}
\usepackage[pagebackref=true,breaklinks=true,colorlinks=true,citecolor = blue,bookmarks=false]{hyperref}

\newtheorem{definition}{Definition}
\newtheorem{theorem}{Theorem}

\newtheorem{remark}{Remark}

\newtheorem{proposition}{Proposition}
\newtheorem{corollary}{Corollary}

\def\BibTeX{{\rm B\kern-.05em{\sc i\kern-.025em b}\kern-.08em
    T\kern-.1667em\lower.7ex\hbox{E}\kern-.125emX}}
\begin{document}

\title{Unsupervised Change Detection using DRE-CUSUM\\
\thanks{This work has been supported in part by NSF Grants CAREER 1651492,
CNS 1715947, CCF 2100013 and the 2018 Keysight Early Career Professor
Award.}
}

\author{
Sudarshan Adiga \qquad Ravi~Tandon\\ 
Department of Electrical and Computer Engineering\\
University of Arizona, Tucson, AZ, USA\\
E-mail: \{\textit{adiga, tandonr}\}@email.arizona.edu
}

\maketitle

\begin{abstract}
This paper presents DRE-CUSUM, an unsupervised density-ratio estimation (DRE) based approach to determine statistical changes in time-series data when no knowledge of the pre-and post-change distributions are available. The core idea behind the proposed approach is to split the time-series at an arbitrary point and estimate the ratio of densities of distribution (using a parametric model such as a neural network) before and after the split point. The DRE-CUSUM change detection statistic is then derived from the cumulative sum (CUSUM) of the logarithm of the estimated density ratio. We present a theoretical justification as well as accuracy guarantees which show that the proposed statistic can reliably detect statistical changes, irrespective of the split point. While there have been prior works on using density ratio based methods for change detection, to the best of our knowledge, this is the first unsupervised change detection approach with a theoretical justification and accuracy guarantees. The simplicity of the proposed framework makes it readily applicable in various practical settings (including high-dimensional time-series data); we also discuss generalizations for online change detection. We experimentally show the superiority of DRE-CUSUM using both synthetic and real-world datasets over existing state-of-the-art unsupervised algorithms (such as Bayesian online change detection, its variants as well as several other heuristic methods).
\end{abstract}


\section{Introduction}
\label{sec: Introduction}

Change detection is the process of identifying deviations in the statistical behavior of time series data, and finds numerous applications, such as detection of distributed denial of service (DDoS) attacks \cite{cardenas2004distributed}, real-time surveillance \cite{huwer2000adaptive}, video segmentation \cite{cotsaces2006video,huang2001robust}, event prediction \cite{steyvers2006prediction,wong2005s} and healthcare monitoring \cite{van2006model,mckeown1998new}. 
To describe the canonical problem of change detection, let us consider a time-series data, denoted by $X_{[1:n]}\triangleq (x_1, x_2, \ldots x_n)$ with a single change point at some unknown time $T^{*}$.
Elements of the sub-sequence $X_{[1:T^*-1]}$ are i.i.d. and sampled from a distribution $P_1$, whereas the elements of sub-sequence $X_{[T^{*}:n]}$ are sampled from a distribution $P_2$. 
The goal of offline change detection is to efficiently determine $T^{*}$ \cite{basseville1993detection}. 

When the pre- and post- change distributions $P_1$, and $P_2$ are known, one can obtain the maximum-likelihood (ML) estimate for the change point using cumulative-sum (CUSUM) of log-likelihood ratios based statistic \cite{wald1948optimum, page1954continuous} (denoted as $S_k = \sum_{t=0}^{k} \log\left({P_2(x_t)}/{P_1(x_t)}\right)$). The main intuition behind CUSUM statistic stems from the expected values of the log-likelihood ratio $P_2(.)/P_1(.)$, before and after $T^{*}$, which is,
\begin{align}
    {\mathbb{E}_{x_t}}\log \left(\frac{P_2(x_t)}{P_1(x_t)}\right) =
    \begin{cases}
        -KL(P_1||P_2), & t < T^{*} \\
        KL\left({P}_2||{P}_1\right), & t \geq T^{*}
    \end{cases}
\end{align}

Since Kullback-Leibler (KL) divergence is non-negative, the CUSUM statistic has a negative expected slope for any $t < T^{*}$, and conversely, positive expected slope for $t \geq T^{*}$.
However, the limitation of the ML-and CUSUM approaches is that they can be applied only when $P_2(x)/P_1(x)$ can be accurately computed for any $x$.
Moreover, in several real-world applications the distributions before and after the change point (denoted by $P_1, P_2$, respectively) are unknown  \cite{hawkins2010nonparametric}, and hence these approaches are impracticable.

\noindent \underline{\textbf{Main Contributions:}} 
In this work, we focus on the challenging setting for change detection when we have no knowledge about pre-and post-change distributions. 
We do not make any assumptions on the underlying probability distributions, i.e., we consider a non-parametric setting. 
The core idea of our proposed methodology is as follows: suppose we observe a time series $X_{[1:n]}$ with an unknown change point at $T^{*}$.
We split the time-series data at an arbitrarily chosen time $T_{\text{split}}$ (say $n/2$) to obtain two sub-sequences as $X_{[1:{T_{\text{split}}-1}]}\sim P_{\text{left}}$, and  $X_{[T_{\text{split}}:n]} \sim P_{\text{right}}$.
We then propose DRE-CUSUM, an unsupervised change detection statistic which \textit{mimics} the conventional CUSUM statistic, with the difference that $P_2(x)/P_1(x)$ is replaced by the estimate of the density ratio $P_{\text{left}}(x)/P_{\text{right}}(x)$. 
In Proposition \ref{thm: Theorem-1}, we prove the surprising result that the DRE-CUSUM statistic possesses theoretical properties analogous to the conventional CUSUM statistic, by showing that 
\begin{align}
    {\mathbb{E}_{x_t}}\log \left(\frac{P_{\text{left}}(x_t)}{P_{\text{right}}(x_t)}\right) =
    \begin{cases}
        > 0, & \text{for } t < T^{*} \\
        < 0, & \text{for } t \geq T^{*}
    \end{cases}
    \label{eq: main-result}
\end{align}
\textit{The highlight of \eqref{eq: main-result} is the fact that it always holds true irrespective of the choice of $T_{\text{split}}$}.
In addition, we also prove accuracy guarantees for DRE-CUSUM by determining the bounds on the probability of error of the estimated change point given that the estimator can correctly compute the density ratio with high probability. 

Furthermore, the theoretical results supporting the use of DRE-CUSUM statistic for unsupervised change detection do not make any assumptions on the density ratio estimators. Therefore, in practice, one can leverage and choose from a wide variety of density ratio estimation techniques \cite{sugiyama2008direct, kanamori2009least} to estimate $P_{\text{left}}(.)/P_{\text{right}}(.)$. This allows a quite general and efficient framework for unsupervised change detection applicable for high-dimensional data. 

We then present generalizations of the DRE-CUSUM approach for detecting multiple changes as well as for online-change detection. We also discuss the possible failure modes of the proposed approach and some possible methods to overcome them. 
Lastly, we present a comprehensive set of experimental results\footnote{Source code for experiments available at: \newline \url{https://www.dropbox.com/sh/xkchf2iajge57jq/AACztXuP-W16VSkUuoTcNY0ya?dl=0}} to demonstrate the superiority in performance of DRE-CUSUM over other change detection methods using both synthetic and real-world datasets.

\noindent \textbf{\underline{Related work:}} 
Several existing algorithms such as sequential probability ratio test (SPRT), generalized likelihood ratio test (GLRT), CUSUM and its variants such as weighted CUSUM are based on the assumption that the density ratios can be readily computed for devising test-statistics for change detection \cite{basseville1993detection}.
To overcome this, one can estimate the distributions using Kernel density estimation (KDE) and histogram-based approach \cite{izenman1991review,sebastiao2008monitoring}. 
Density estimation techniques for change detection has several shortcomings. 
Firstly, density estimation based approaches are limited to applications with low-dimensional observations \cite{li2014density}.
Secondly, the performance of change detection using density estimation techniques could deteriorate when division by the estimated density magnifies the estimation error of the log-likelihood ratio \cite{sugiyama2012density}.
Lastly, the samples from pre- and post-change distributions must be labelled. 

\noindent An alternate technique to determine the change points is by directly estimating the density ratios \cite{sugiyama2008direct, kanamori2009least, nam2015direct, vapnik2013constructive}, an overview which has been presented in Appendix \ref{sec: Methods for Estimating Likelihood Ratios}.
An unsupervised change detection algorithm using a DRE framework has been proposed in \cite{hushchyn2021generalization}. 
However, no theoretical justification and accuracy guarantees have been provided for this approach. 
Clustering, and Bayesian change detection (BCD) are some of the other techniques that do not require explicit labelling of pre- and post- change samples \cite{zakaria2012clustering, adams2007bayesian, fearnhead2006exact}.
However, such approaches often lack a statistical justification and are dependent on ordering of data as well as the initial parameters \cite{luppino2017clustering}.
Furthermore,  BCD based methods \cite{adams2007bayesian, fearnhead2006exact} make assumptions on the underlying data generating distributions and on the change points (also referred to as data, and change-point priors, respectively).
Other limitations of the BCD based methods are that: (a) these approaches are suitable for low-dimensional data and are difficult to implement for high-dimensional data, and (b) are prone to high false alarm rates.

There are several other heuristic approaches for unsupervised change detection.  For instance, dynamic programming techniques can be used for change detection by performing an ordered search to determine the best possible set of partitions, given the total number of change points apriori \cite{truong2020selective}.  
Linearlized penalty segmentation (Pelt) overcomes the need for the number of change points by minimizing a penalty term that is itself a function of the estimated change points \cite{killick2012optimal,wambui2015power}. These methods however, have a higher computational cost. Binary segmentation (BinSeg) finds an approximate solution by first finding the best instance (first change point estimate) to split the time-series into two sub-sequences \cite{fryzlewicz2014wild}.  This operation is then executed in the derived sub-sequences in a iterative manner to determine all possible change points.  The main drawback of this approach is that errors in the initial stages of the algorithm can propagate for subsequent steps. 
An adaptive search method to find the change point by maximizing a CUSUM based statistic when the pre-and post-change distributions are from Gaussian distribution has been proposed in \cite{kovacs2020optimistic}.

In Section \ref{sec: experiments}, we compare the performance of the proposed DRE-CUSUM technique with its unsupervised change detection counterparts such as BCD and its variants, as well as several other heuristic approaches as discussed above.

\section{Unsupervised Change Detection}
\label{sec: DRE-CUSUM}
\noindent \underline{\textbf{Problem formulation}}-
\textit{\noindent Consider the time-series data $X_{[1:n]}$ which undergoes a change at an unknown time $T^{*}$. 
The samples $X_{[1:T^{*}-1]}$ are i.i.d., drawn from a distribution $P_1$, and  samples $X_{[T^{*}:n]}$ are i.i.d., drawn from a distribution $P_2$. 
The goal of unsupervised offline change detection is to estimate $T^{*}$, when $P_1$, $P_2$ are unknown. We discuss the online setting later in this section.}

 \noindent When pre-and post-change distributions are known, one can obtain the change point estimate ($\hat{T}_{ML}$) using maximum likelihood (ML) \cite{basseville1993detection}:
\begin{align}
\hat{T}_{ML} = \argmax_{t}  \sum_{i=t}^{n} \log \left(\frac{P_2(x_{i})}{P_1(x_{i})}\right)
\label{eq: max-likelihood}
\end{align}
\noindent However, the ML approach can be applied if either the distributions $P_1$ and $P_2$ are known, or the density ratio $P_2/P_1$ can be accurately computed.
The need for the information on the distributions and their corresponding order in the time series makes the ML approach infeasible for most change detection applications.

\begin{figure}[!t]
\centering
\begin{subfigure}[b]{0.46\textwidth}
   \includegraphics[width=1\linewidth]{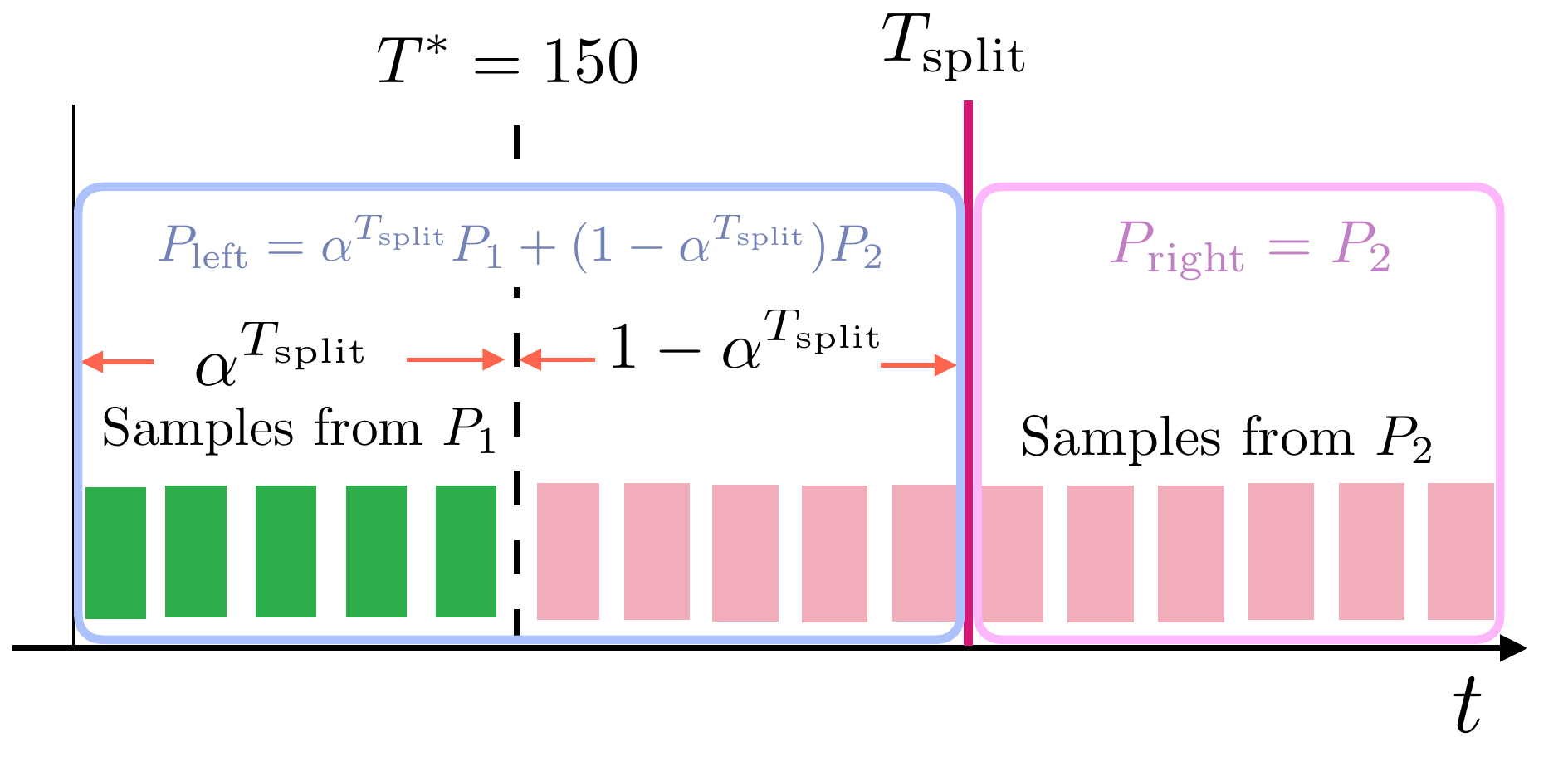}
   \caption{Time-series data with a single change point at $T^{*}$, when split at $T_{\text{split}} \geq T^{*}$ yields two distributions: $P_{\text{left}} = \alpha^{T_{\text{split}}} P_1 + (1-\alpha^{T_{\text{split}}}) P_2$, and $P_{\text{right}} = P_2$. Alternatively, when $T_{\text{split}} \leq T^{*}$, we have $P_{\text{left}} = P_1$, and $P_{\text{right}} = \alpha^{T_{\text{split}}} P_1 + (1-\alpha^{T_{\text{split}}})P_2$.}
    \label{fig: dr-oracle-cusum-mean-change-a} 
\end{subfigure}

\begin{subfigure}[b]{0.46\textwidth}
   \includegraphics[width=1\linewidth]{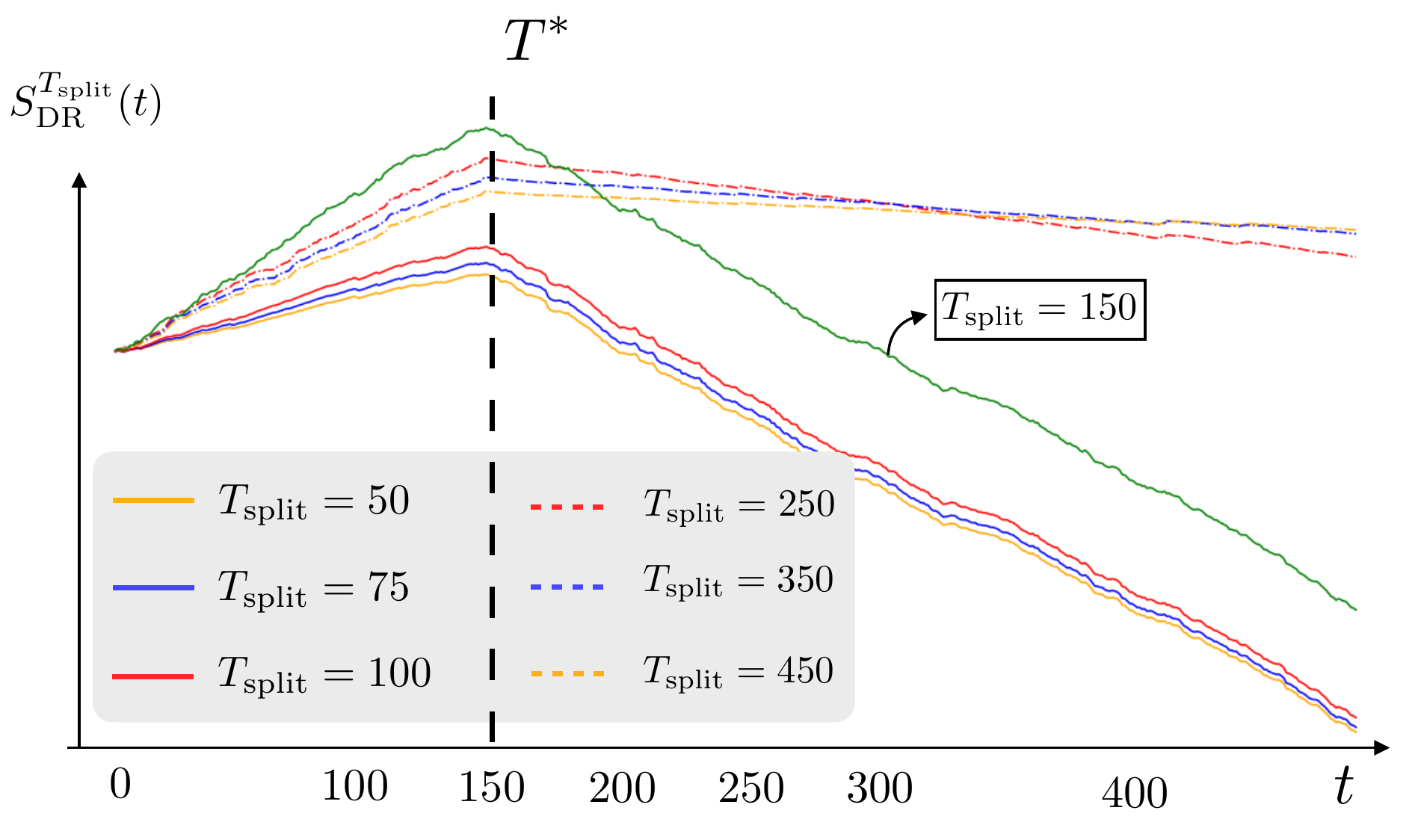}
   \caption{{Plot of density-ratio based CUSUM statistic $S^{T_{\text{split}}}_{\text{DR}}(t)$ vs $t$ for $10$-dimensional time-series with $500$ samples, and an unknown change point $T^{*} = 150$, such that $X_{[1:149]} \sim P_1 = \mathcal{N}(\vec{\mu}_1,I)$, and $X_{[150:500]} \sim P_2 = \mathcal{N}(\vec{\mu}_2,I)$, where entries of the mean vectors $\vec{\mu}_1$,$\vec{\mu}_2$ are sampled from $\text{Unif.}[0.0,0.4]$, $\text{Unif.}[0.6,1.0]$, respectively.}} 
    \label{fig: dr-oracle-cusum-mean-change-b}
\end{subfigure}

\caption{Unsupervised change detection statistic $S^{T_{\text{split}}}_{\text{DR}}(t)$ for different $T_{\text{split}}$ values. From Fig. \ref{fig: dr-oracle-cusum-mean-change-b}, we observe that the slope of $S^{T_{\text{split}}}_{\text{DR}}(t)$ changes at $T^{*}$ irrespective of the value of $T_{\text{split}}$.} 
\vspace{-10pt}
\end{figure}

\noindent To alleviate the disadvantages of the ML approach, we consider a setting in which we do not know the pre-and post-change distributions. 
Given the time-series $X_{[1:n]}$, we split the time-series at a point $T_{\text{split}}$, to obtain two sub-sequences: $ X_{[1:T_{\text{split}}-1]} \sim P_{\text{left}}$, 
$X_{[T_{\text{split}}:n]} \sim P_{\text{right}}$, where $P_{\text{left}}$ and $P_{\text{right}}$ denote the corresponding distributions of the samples (see Fig. \ref{fig: dr-oracle-cusum-mean-change-a}). 
Based on the relative position of $T_{\text{split}}$ with respect to $T^{*}$, either $P_{\text{left}}$ or $P_{\text{right}}$ is a mixture distribution, and conversely, the other is a pure distribution (either $P_1$ or $P_2$). We next define the density-ratio (DR) based cumulative-sum (CUSUM) of likelihood ratio based statistic $S^{T_{\text{split}}}_{\text{DR}}(t)$, $\forall t \in [1,n]$.
\begin{align}
    S^{T_{\text{split}}}_{\text{DR}}(t) \triangleq \sum_{j=1}^{t} \log\left(\frac{P_{\text{left}}(x_j)}{P_{\text{right}}(x_j)}\right).
\end{align}

\noindent Fig. \ref{fig: dr-oracle-cusum-mean-change-b} depicts $S^{T_{\text{split}}}_{\text{DR}}(t)$, $\forall t \in [1,n]$ for different values of $T_{\text{split}}$ (i.e. both $T_{\text{split}} \geq T^{*}$ and $T_{\text{split}} < T^{*}$) for a 10-dimensional multivariate Gaussian time-series undergoing a mean change at $T^{*}=150$. As seen from this example, the change point $T^{*}$ manifests itself in $S^{T_{\text{split}}}_{\text{DR}}(t)$ through a slope change at $T^{*}$, \textit{irrespective} of the choice of $T_\text{split}$.
Furthermore, we note that $S^{T_{\text{split}}}_{\text{DR}}(t)$ for $T_{\text{split}} = T^{*}$ corresponds to the maximum likelihood-estimate in \eqref{eq: max-likelihood}.
We can make the following observations about the nature of $S^{T_{\text{split}}}_{\text{DR}}(t)$: (a) the slope before the change point $T^{*}$ is always positive, (b) the slope after $T^{*}$ is always negative. 
This intuition is formalized in our first result which is stated in Proposition \ref{thm: Theorem-1}.
Subsequently, we use Proposition \ref{thm: Theorem-1} to show that the expected value of the DR-CUSUM statistic $S^{T_{\text{split}}}_{\text{DR}}(t)$ takes its maximum value at the true change point $T^{*}$. In order to present our theoretical results, we define a mixture distribution as well as a non-negative function composed of weighted KL divergence.

\begin{definition}
Given two distributions $P_1$, $P_2$, we define a parametric mixture distribution (for any $0\leq \lambda\leq 1$) as 
\begin{align}
P{(\lambda)} = \lambda P_1 + (1-\lambda) P_2    
\end{align}
Furthermore, for any $\lambda \in (0,1), \gamma \in [0,1]$, we define the following non-negative function $f_{i}(\gamma,\lambda)$ for $i = \{1,2\}$ as follows:
\begin{align*}
f_{i}(\gamma,\lambda)\overset{\Delta}{=}\frac{1}{\gamma}KL(P{(\lambda)}||P_{i}) + \frac{1-\gamma}{\gamma}KL(P_{i}||P{(\lambda)})
\end{align*}
\end{definition}

\noindent Our first result is stated in the following Proposition. Proof of Proposition \ref{thm: Theorem-1} is presented in Appendix \ref{sec:Theorem-1-Proof}.

\begin{proposition}
\label{thm: Theorem-1}
If $T_{\text{split}} \leq T^{*}$, then
\begin{align*}
 {\mathbb{E}_{x_{t}}\left[\log\frac{P_{\text{left}}(x_{t})}{P_{\text{right}}(x_{t})}\right]}=
\begin{cases}
 KL(P_1||P(1-\alpha_1)), & t \in [1,T^{*}) \\
-f_{1}(\alpha_1,1-\alpha_1), & t \in [T^{*},n]
\end{cases}
\end{align*}
If $T_{\text{split}} \geq T^{*}$, then
\begin{align*}
 {\mathbb{E}_{x_{t}}\left[\log\frac{P_{\text{left}}(x_t)}{P_{\text{right}}(x_t)}\right]}=
\begin{cases}
 f_{2}(\alpha_2,\alpha_2), & t \in [1,T^{*}) \\
-KL(P_2||P(\alpha_2)), &   t \in [T^{*},n]
\end{cases}
\end{align*}
where, $\alpha_1 = \frac{n-T^{*}}{n-T_{\text{split}}}$ and $\alpha_2 = \frac{T^{*}}{T_{\text{split}}}$.
\end{proposition}
\noindent We next present Corollary \ref{cor: Corollary-1}, which is a direct consequence of the result in Proposition \ref{thm: Theorem-1}.
 \begin{corollary}
 \label{cor: Corollary-1}
 The DR-CUSUM statistic satisfies
 \begin{align}
     T^{*} = \argmax_{t} \mathbb{E}\left[S^{T_{\text{split}}}_{\text{DR}}(t)\right] \text{ }\forall T_{\text{split}}
 \end{align}
 \end{corollary}
\noindent The key significance of the above result is that the expected value of the DR-CUSUM statistic takes it's maximum value at the uknown change point $T^{*}$, irrespective of the split point $T_{\text{split}}$. This result provides the motivation for the density ratio based estimator, denoted by $\hat{T}_{\text{DR-CUSUM}}$ for unsupervised change detection :
\begin{align}
    \hat{T}_{\text{DR-CUSUM}} = \argmax_{t} S^{T_{\text{split}}}_{\text{DR}}(t) 
    \label{eq: meta-algorithm}
\end{align}
In order to quantify the performance of the change point estimate $\hat{T}_{\text{DR-CUSUM}}$, we adopt the notion of $(\alpha, \beta)$-accuracy as introduced in \cite{cummings2018differentially}, and defined next. 
\begin{definition}  
A change point estimate $\hat{T}$ is $(\alpha,\beta)-$ accurate for a change point $T^{*}$ if for $\alpha, \beta \in [0,1]$-
\begin{align}
    P[|\hat{T} - T^{*}| < \alpha] \geq 1-\beta. 
\end{align}
\end{definition}

We next present our second main result, which provides $(\alpha, \beta)$-accuracy guarantees for the density ratio based estimator $\hat{T}_{\text{DR-CUSUM}}$, under the assumption that the log-likehood ratio $\log(P_\text{left}(.)/P_\text{right}(.))$ is bounded by a constant $A$.

\begin{theorem}
\label{thm: Theorem-2}
For any $0< \beta<1$, $\hat{T}_{\text{DR-CUSUM}}$ satisfies $(\alpha, \beta)$-accuracy for 
\begin{align}
    \alpha = \frac{2A^2}{C^2} \log \left( \frac{32}{3 \beta} \right),
\end{align}
where the constant $C$ represents the minimum value of the expected log-likelihood ratio $\frac{P_{\text{left}}(x)}{P_{\text{right}}(x)}$ before and after the change point $T^{*}$, and is given as follows
\begin{align*}
&C =\\
&\begin{cases}
\min \left( KL(P_1||P(1-\alpha_1)),   f_{1}(\alpha_1,1-\alpha_1)\right),  T_{\text{split}} \leq T^{*}\\
\min \left( f_{2}(\alpha_2,\alpha_2),  KL(P_2||P(\alpha_2))\right),  T_{\text{split}} \geq T^{*},
\end{cases}
\end{align*}
and $\alpha_1 = \frac{n-T^{*}}{n-T_{\text{split}}}$ and $\alpha_2 = \frac{T^{*}}{T_{\text{split}}}$.
\end{theorem}
\noindent Proof of Theorem \ref{thm: Theorem-2} is presented in Appendix \ref{sec:Theorem-2-Proof}.  
The main idea behind the proof of Theorem \ref{thm: Theorem-2} follows from considering the upper bound of the probability with which DR-CUSUM statistic assumes its maximum value at time instances at least $\alpha$ away from the true change point $T^{*}$.
This can then be expressed in terms of the probability with which sum of the log-likelihood ratios of data points in the time-series between $T^{*}$ and all other time instances at distance  $\alpha$ from $T^{*}$ is greater than $0$.
By subtracting the expected value of the log-likelihood ratio (i.e., KL divergence), this simplifies to the sum of zero mean i.i.d random variables to which we apply Ottaviani’s inequality \cite{van1996weak} to obtain the ($\alpha$,$\beta$) accuracy guarantees for the density ratio based estimator $\hat{T}_{\text{DR-CUSUM}}$. 

\begin{remark} \textit{Dependence on KL divergence:} For fixed $A$, $\beta$, we note that $\alpha$ is inversely proportional to $C^2$ which is related to the KL-divergence between pure and mixture distributions (subject to $T_{\text{split}}$), and this implies that the larger the change in distributions before and after $T^{*}$, higher is the accuracy of $\hat{T}_{\text{DR-CUSUM}}$.
\end{remark}


\subsection{The DRE-CUSUM Estimator}
\label{sec: Unsupervised-CD-DRE-CUSUM}
\noindent We now present the DRE-CUSUM estimator $\hat{T}_{\text{DRE-CUSUM}}$. Specifically, we split the time series at $T_{\text{split}}$ and compute the statistic $S^{T_{\text{split}}}_{\text{DRE}}$ as follows:
\begin{align}
    S^{T_{\text{split}}}_{\text{DRE}}(t) = \sum_{i=1}^{t} \log \left( \hat{w}(x) \right),
    \label{eq: DRE-CUSUM-dre}
\end{align}
where $\hat{w}(x)$ is an estimate of the density ratio ${P_{\text{left}}(x)}/{P_{\text{right}}(x)}$ which is obtained by density ratio estimation (DRE) models using samples from distributions $P_{\text{left}}$ and $P_{\text{right}}$.
The DRE-CUSUM estimator $\hat{T}_{\text{DRE-CUSUM}}$ is then obtained as follows:
\begin{align}
    \hat{T}_{\text{DRE-CUSUM}} = \argmax_{t} S^{T_{\text{split}}}_{\text{DRE}}(t)
    \label{eq: DRE-CUSUM-estimate}
\end{align}

\noindent We label the algorithm using the DRE for unsupervised change detection as DRE-CUSUM, whose steps are described in Algorithm \ref{alg: Algorithm_CD_1}.
Assuming that the DRE models correctly estimate the density ratio with high probability \cite{vapnik1999overview,stefanyuk1986estimation}, then the Proposition \ref{thm: Theorem-1} and Theorem \ref{thm: Theorem-2} extend to $S^{T_{\text{split}}}_{\text{DRE}}(t)$ and $\hat{T}_{\text{DRE-CUSUM}}$, respectively.

\begin{algorithm}[!h]
	\caption{Unsupervised Single Change Point Detection using DRE-CUSUM.} 
	\label{alg: Algorithm_CD_1} 
	\begin{algorithmic}
		\STATE{ Input time-series data: $\left(x_{1}, x_{2}, ..,x_{T^{*}},...,x_{n}\right)$}
		\STATE{\underline{\textbf{{1. Density Ratio Estimator (DRE) Training}}}\vspace{0.1cm}}
		\STATE{Divide the time-series data at $T_\text{{split}}$ (say $T_\text{{split}} = \frac{n}{2}$) to obtain- (i) $X_{[1:{T_\text{split}-1}]} \sim P_{\text{left}}$, (ii) $X_{[{T_\text{split}}:n]} \sim P_{\text{right}}$.}
		\FOR{number of epochs}
		\STATE{a. Sample $N_1$,$N_2$ samples from $P_{\text{left}}$, $P_{\text{right}}$, respectively.}
		\STATE{b. Train DRE to determine $\hat{w}(x)$, an estimate of the density ratio ${P_{\text{left}}(x)}/{P_{\text{right}}(x)}$.
		(see Appendix \ref{sec: Methods for Estimating Likelihood Ratios}.)
		}
		\ENDFOR
		
		\STATE{\underline{\textbf{{2. DRE-CUSUM based Change Detection}}}\vspace{0.1cm}}
		\STATE{a. Compute $S^{T_{\text{split}}}_{\text{DRE}}(t) = \sum_{j=1}^{t} \log\left(\hat{w}(x_j)\right) $}
		\STATE{b.  List the time instance $\hat{T}$ (estimated change point) at which there is a change in slope. 
		}
		\STATE{\underline{\textbf{{3. Verification Step}}}}
		\STATE{Repeat steps 1,2 setting $T^{'}_\text{{split}} = \hat{T}$ (but not equal to $\frac{n}{2}$), and find $\hat{T}_{\text{DRE-CUSUM}} = \argmax_{t}S^{T'_{\text{split}}}_{\text{DRE}}(t)$.
		Verify that  $\hat{T} = \hat{T}_{\text{DRE-CUSUM}}$ is the only slope change in $S^{T'_{\text{split}}}_{\text{DRE}}(t)$.}
	\end{algorithmic}
\end{algorithm}

\begin{remark}
Role and Impact of DRE estimation models: we highlight that the proposed approach can in principle rely upon any of the existing methods (such as \cite{sugiyama2008direct, kanamori2009least}) for density ratio estimation given the samples from $P_{\text{left}}$ and $P_{\text{right}}$. We study the impact of the choice of DRE estimation on the change detection performance in Section \ref{sec: experiments}. 
\end{remark}

\begin{remark}
Justification of the verification Step in Algorithm $1$: When implementing DRE-CUSUM on real-world data, it is possible to observe two slope changes: one at $T_{\text{split}}$ (the initial split point) and another at the potential change point. 
Therefore, we first list all possible slope changes in $S^{T_{\text{split}}}_{\text{DRE}}(t)$ as mentioned in step 2 of Algorithm \ref{alg: Algorithm_CD_1}.
In order to correctly declare the change point, we propose a verification step: repeat steps 1,2 in Algorithm \ref{alg: Algorithm_CD_1} by setting
$T^{'}_{\text{split}}$ equal to the all instances in this list; and find $\hat{T}_{\text{DRE-CUSUM}} = \argmax_{t}S^{T'_{\text{split}}}_{\text{DRE}}(t)$.
If $\hat{T}_{\text{DRE-CUSUM}} = T^{*}$, we expect to observe only one slope change in $S^{T^{'}_{\text{split}}}_{\text{DRE}}(t)$ at $T^{*}$, and thereby allowing us to rule out $T_{\text{split}}$ as a change point. 
\end{remark}

\begin{remark}
Robustness of DRE-CUSUM to $T_{\text{split}}$ and the choice of $T_{\text{split}}$: a natural question to ask is the following: how sensitive is the DRE-CUSUM estimator to the choice of $T_{\text{split}}$? We present experimental results in Section \ref{sec: experiments} which show the impact of choice of $T_{\text{split}}$ and the distance $\lvert T_{\text{split}} - T^{*} \rvert$ between the split point and the unknown change point. 
The second question to ask is: what is the natural choice of $T_{\text{split}}$?
For sufficiently large number of samples from $P_{\text{left}}$ and $P_{\text{right}}$, \cite{vapnik2013constructive} shows that $\hat{w}(x)$ converges in probability to the true density ratio $P_{\text{left}}(x)/P_{\text{right}}(x)$.
This provides an intuitive explanation behind the choice of $T_{\text{split}} = \frac{n}{2}$ in Algorithm \ref{alg: Algorithm_CD_1}.

\end{remark}

\subsection{Generalizations of DRE-CUSUM}
\label{sec: generalizations-DRE-CUSUM}
In this section, we discuss generalizations of the DRE-CUSUM algorithm for detecting multiple changes, adaptation for online change detection, as well as some approaches for overcoming potential failure modes. \\
\newline \noindent (a) \textbf{Multiple change detection}
\newline \noindent 
\begin{figure}[!t]
\centering
\begin{subfigure}[b]{0.46\textwidth}
   \includegraphics[width=1\linewidth]{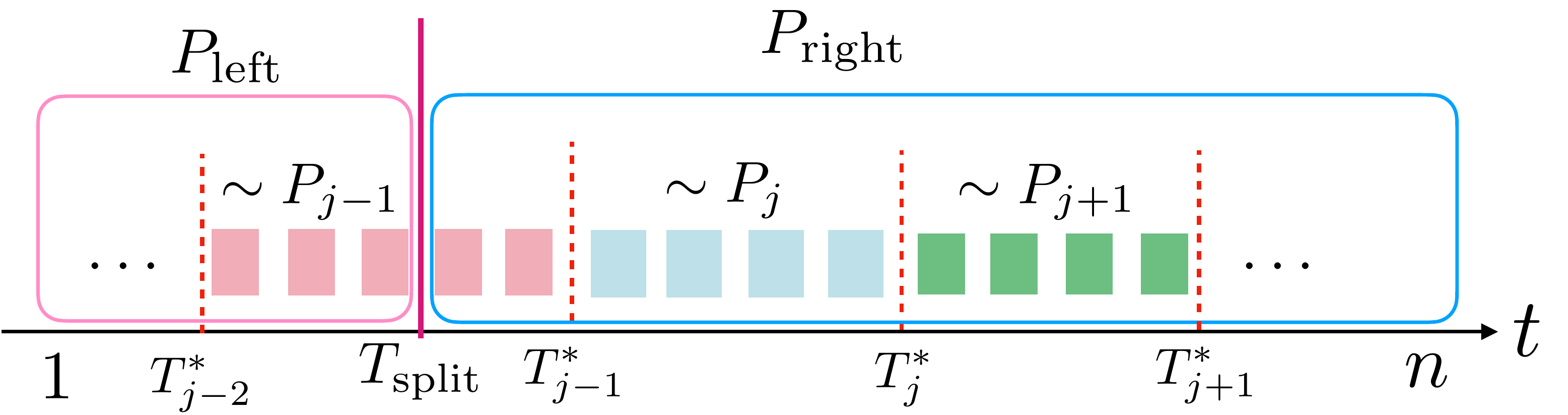}
   \caption{Multiple change point time-series data, where $X_{[T_{j-1}^{*}:T_{j}^{*}]} \sim P_j$.}
   \label{fig:DRE-CUSUM-mcp-a} 
\end{subfigure}

\begin{subfigure}[b]{0.46\textwidth}
   \includegraphics[width=1\linewidth]{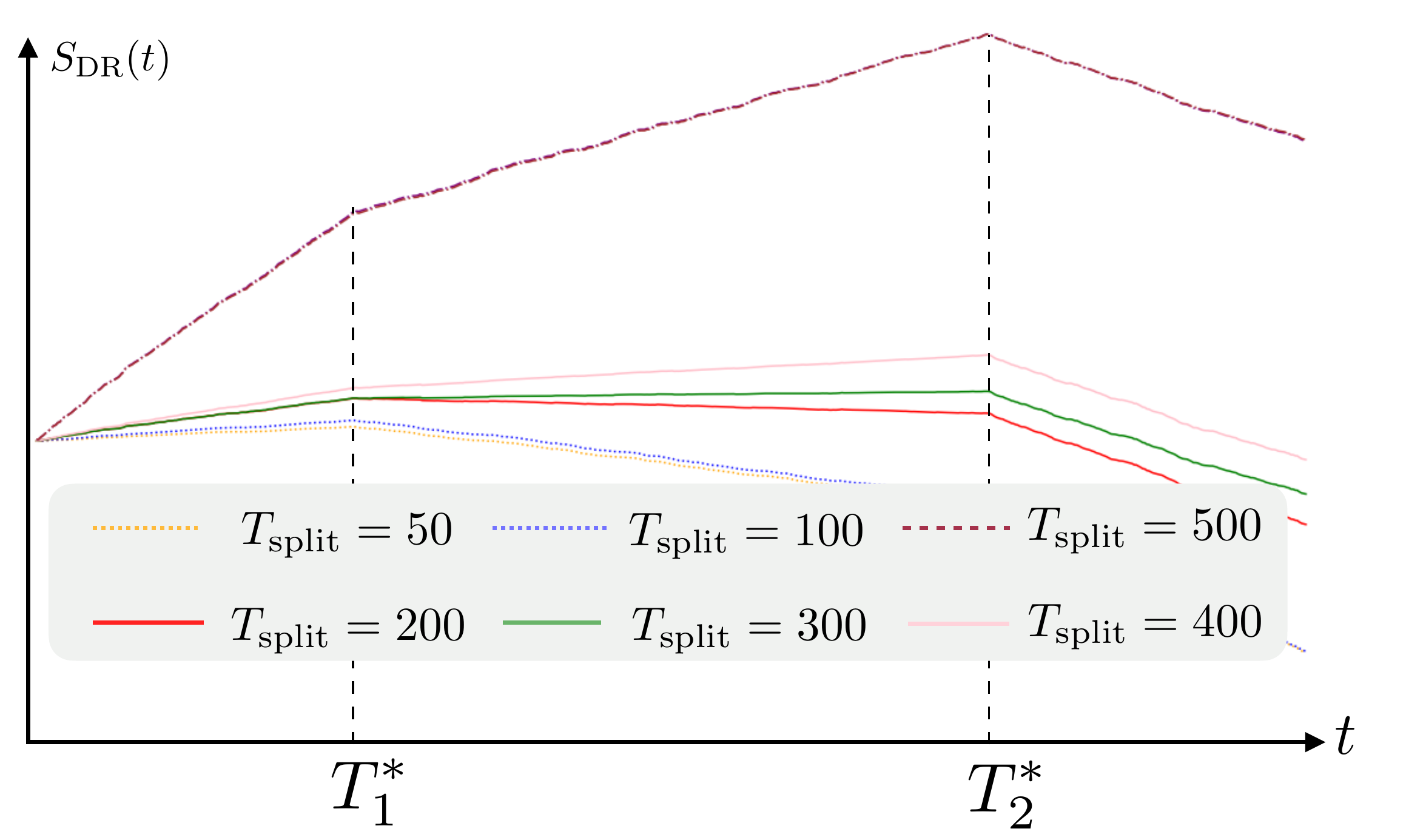}
   \caption{{$S^{T_{\text{split}}}_{\text{DR}}(t)$ vs $t$ for $10$-dimensional time-series of length $600$ with two change points  $T_1^{*} = 150$,  $T_2^{*} = 450$. $X_{[1:149]}$, $X_{[150:449]}$ and $X_{[450:599]}$ follow multivariate gaussian distributions with mean vectors are sampled from $\text{Unif.}[0,0.4]$, $\text{Unif.}[0.6,1.0]$, and $\text{Unif.}[1.6,2.0]$, respectively, and identity covariance matrix.
   }} 
   \label{fig:DRE-CUSUM-mcp-b} 
\end{subfigure}
\caption{Unsupervised multiple change detection statistic $S^{T_{\text{split}}}_{\text{DR}}(t)$ for different $T_{\text{split}}$ values.}
\vspace{-10pt}
\end{figure}

Consider the time series $X_{[1:n]}$ with $K\geq 1$ change points, denoted as $T^{*}_{1}\leq  T^{*}_{2}\leq \cdots \leq T^{*}_{K}$. The sub-sequence $X_{[T_{j-1}^{*} , T_j^{*}]}$ in the $j$th segment is i.i.d. with samples drawn from an unknown distribution  $P_j$ for $j=1, 2,\ldots K$ (see Fig. \ref{fig:DRE-CUSUM-mcp-a}).  We show that a similar approach of splitting the time-series followed by computing the DRE-CUSUM statistic can be leveraged for detecting more than one change points. To provide the intuition behind this, consider any split point $T_{split}$, and as before, suppose that we can compute the ratio $P_{\text{left}}(x)/P_{\text{right}}(x)$. Analogous to Proposition \ref{thm: Theorem-1}, it can be readily shown that for every $t \in [T_{j-1}^{*} , T_j^{*}]$, the expected value of the $\log(\cdot)$ of the density ratio is given as:
\begin{align}
\mathbb{E}_{x_t} \left[\log \frac{P_{\text{left}}(x_t)}{P_{\text{right}}(x_t)}\right] = \underbrace{KL(P_j||P_{\text{right}}) - KL(P_j||P_{\text{left}})}_{=\Delta_{j}}
\label{eq: DRE-CUSUM-mcp} \nonumber
\end{align}
As discussed in the previous section, the slope of the DRE-CUSUM statistic will be proportional to the quantity $\Delta_j$.  Thus, as long as $\Delta_{j} \neq \Delta_{j-1}$ and $\Delta_{j}\neq \Delta_{j+1}$ for all $j=1,2, \ldots, K$, we can expect distinct slopes in the DRE-CUSUM statistic for each segment in the time-series. In Fig. \ref{fig:DRE-CUSUM-mcp-b}, we show this behaviour for a synthetic 10-dim multivariate Gaussian time-series with two change points. The instances of the slope change are potential candidates for the estimated change points.  In Section \ref{sec: experiments}, we also provide a comprehensive set of experiments on the use of DRE-CUSUM for detecting multiple change points for a variety of real-world datasets.



\begin{figure}[!t] 
	\vspace{-1pt}
	\centering
	\includegraphics[scale= 0.21]{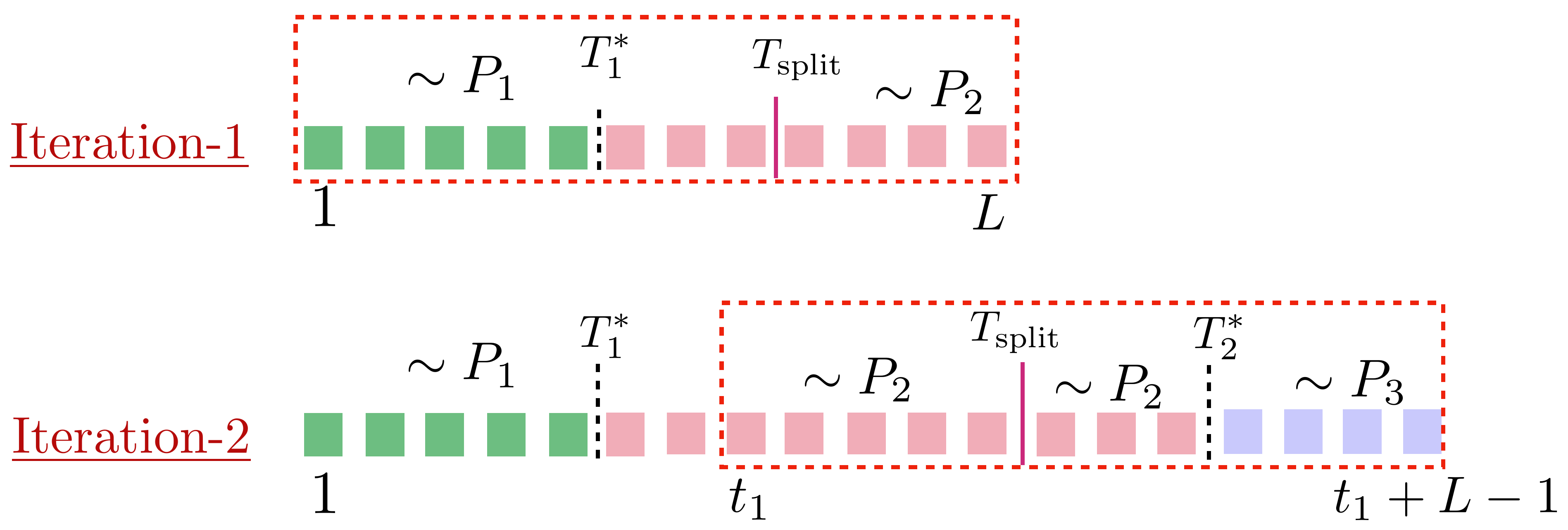}
	\caption{Online adaptation of DRE-CUSUM algorithm. The window size can be either of a fixed size or it can be selected in an adaptive manner based on changes detected in the past.  \label{fig: Online-CD-1}}
\end{figure}

\noindent (b) \textbf{Online change point detection}
\newline \noindent DRE-CUSUM can be readily applied for online change detection by recursively performing Steps 1-3 in Algorithm \ref{alg: Algorithm_CD_1} on real-time data. As shown in Fig. \ref{fig: Online-CD-1}, a simple approach is to consider a window of length $L$ (with $L$ most recent samples collected).
Steps 1-3 in Algorithm \ref{alg: Algorithm_CD_1} can be performed on this window of $L$ samples to determine all change points within this time interval. We slide this window across the time series to consider new observations. A generalization of this approach is to use \textit{adaptive} window sizes depending on past detected changes. Specifically, if we have reliably detected changes in the previous window, then one only needs to keep the most recent samples from the past after the latest detected change point.   
In Section \ref{sec: experiments}, we provide experimental results for DRE-CUSUM for detecting multiple changes in both offline as well as online setting. 

\begin{figure}[!t]
\centering
\begin{subfigure}[b]{0.46\textwidth}
   \includegraphics[width=1\linewidth]{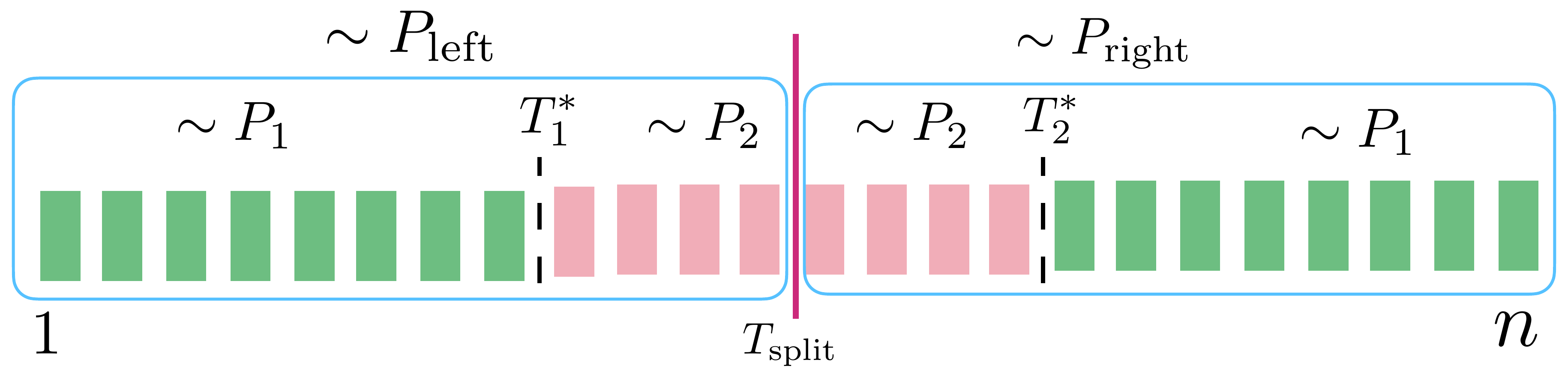}
   \caption{Failure mode: DRE-CUSUM in Algorithm \ref{alg: Algorithm_CD_1} fails to detect the changes $T^{*}_1$, $T^{*}_2$ when $P_{\text{left}} \approx P_{\text{right}}$.}
    \label{fig: Failure-Modes-a} 
\end{subfigure}

\begin{subfigure}[b]{0.46\textwidth}
   \includegraphics[width=1\linewidth]{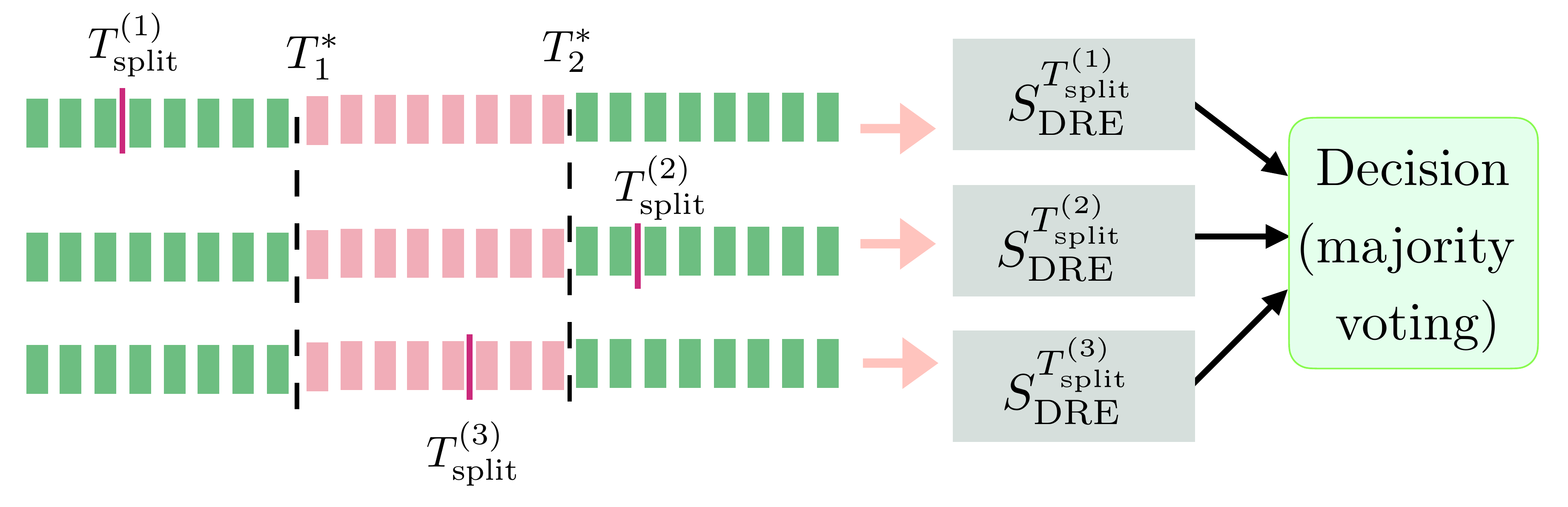}
   \caption{{Proposed approach: Compute DRE-CUSUM statistic for multiple $T_{\text{split}}$} values followed by a combined decision (e.g., majority vote).} 
    \label{fig: Failure-Modes-b} 
\end{subfigure}

\caption{Overcoming failure mode of DRE-CUSUM.}
\end{figure}

\noindent (c) \textbf{Reducing Errors in DRE-CUSUM}\newline
We now discuss how to overcome one of the most common failure modes of the DRE-CUSUM approach using an example as shown in Fig. \ref{fig: Failure-Modes-a}, in which $X_{[1:{T_{\text{split}}-1}]} \sim P_{\text{left}}$, and $X_{[{T_{\text{split}}}:n]} \sim P_{\text{right}}$. If for a $T_{\text{split}}$, it happens that $P_{\text{left}}(x) \approx P_{\text{right}}(x)$, $\forall x$, then 
as a consequence, the KL divergence  $KL(P_{\text{left}} || P_{\text{right}}) \approx 0$. 
In such a scenario, the DRE-CUSUM statistic $S_{\text{DRE}}(t)$ can fail to exhibit a slope change at the unknown change points.  To alleviate this phenomenon, we propose a simple and efficient modification of the Algorithm \ref{alg: Algorithm_CD_1} by considering multiple distinct $T_{\text{split}}$  as shown in the Fig. \ref{fig: Failure-Modes-b}, i.e., we run the DRE-CUSUM algorithm for multiple distinct split points (say $T^{(1)}_{\text{split}}, T^{(2)}_{\text{split}}, \ldots T^{(r)}_{\text{split}}$).  
The change points in the time-series can then be determined by applying a combined decision across the slope changes exhibited by the multiple DRE-CUSUM statistic(s).
Some examples of the combined decision techniques that can be applied here are: (i) majority voting, (ii) weighted sum technique, wherein the weight corresponds to the probability that the slope change at a time instance corresponds to the true change point and is determined by the extent of the slope change.
Furthermore, by using multiple values of $T_{\text{split}}$, we enhance the change detection framework in Algorithm \ref{alg: Algorithm_CD_1} through reduction in the detection errors (i.e. false alarms and mis-detections). 
Another refinement to Algorithm \ref{alg: Algorithm_CD_1} to minimize the errors is by searching for the best $T_{\text{split}}$ according to the proposed adaptive methods in \cite{kovacs2020optimistic}. 
The subsequent $T_{\text{split}}$ can be selected to maximize the value of the statistic $S^{T_{\text{split}}}_{\text{DRE}}$ at time instances with a slope change.

\section{Experiments}
\label{sec: experiments}
In this section, we present a comprehensive set of experiments to show: (i) the robustness of the DRE-CUSUM algorithm, (ii) the superiority of the DRE-CUSUM approach with other unsupervised techniques on both synthetic and real-world datasets, (iii) capability of detecting changes in high-dimensional video datasets.  
Particularly, the experiments on the event detection in video frames highlight the key aspect \textit{ that DRE-CUSUM is capable of demarcating the change points in very high-dimensional time-series data.}


\textit{Performance metrics:} 
For evaluating DRE-CUSUM with other approaches, we use \textit{false alarm rate} (FAR) and \textit{missed detection rate} (MDR) \cite{aminikhanghahi2017survey} which is computed as, 
\begin{align}
    FAR = \frac{FP}{FP+TN} \hspace{1cm} MDR = \frac{FN}{FN+TP}
\end{align}
where,  TP, TN, FP, and FN denote the true positives, true negatives, false positives and false negatives, respectively.

\textit{Architecture of DRE models ($\hat{w}(x)$, estimate of $P_{\text{left}}/P_{\text{right}}$):}
For the scope of the experiments, we consider DRE modeled using kernels and deep neural networks (DNN's). We use the package provided in \cite{pypi2016pypi} for the kernel-based DRE. For the synthetic datasets,  $4$-layered feed-forward neural network based DRE is used with sigmoid, and softplus activations in the hidden, and final layers, respectively. For the change detection on video datasets, we use $4$-layered convolutional neural network, with sigmoid, and softplus activations used in the hidden layers, and final layer, respectively. 
To train a DRE, a wide variety of training objectives such as KLIEP and LSIF have been widely accepted and used \cite{kanamori2009least,sugiyama2008direct}, which we adopt to train the DRE's in the experimental section. 
Details on objective functions to train DRE, input to the DRE and more are presented in Appendix \ref{sec: Methods for Estimating Likelihood Ratios}. 

\subsection{Experiments on synthetic datasets}
We first demonstrate the robustness of DRE-CUSUM to $\lvert T^{*} - T_{\text{split}}\rvert$, and distance between pre-change ($P_1$) and post-change distributions ($P_2$). 

\begin{figure}[!h]
\centering
\begin{subfigure}[b]{0.46\textwidth}
   \includegraphics[width=1\linewidth]{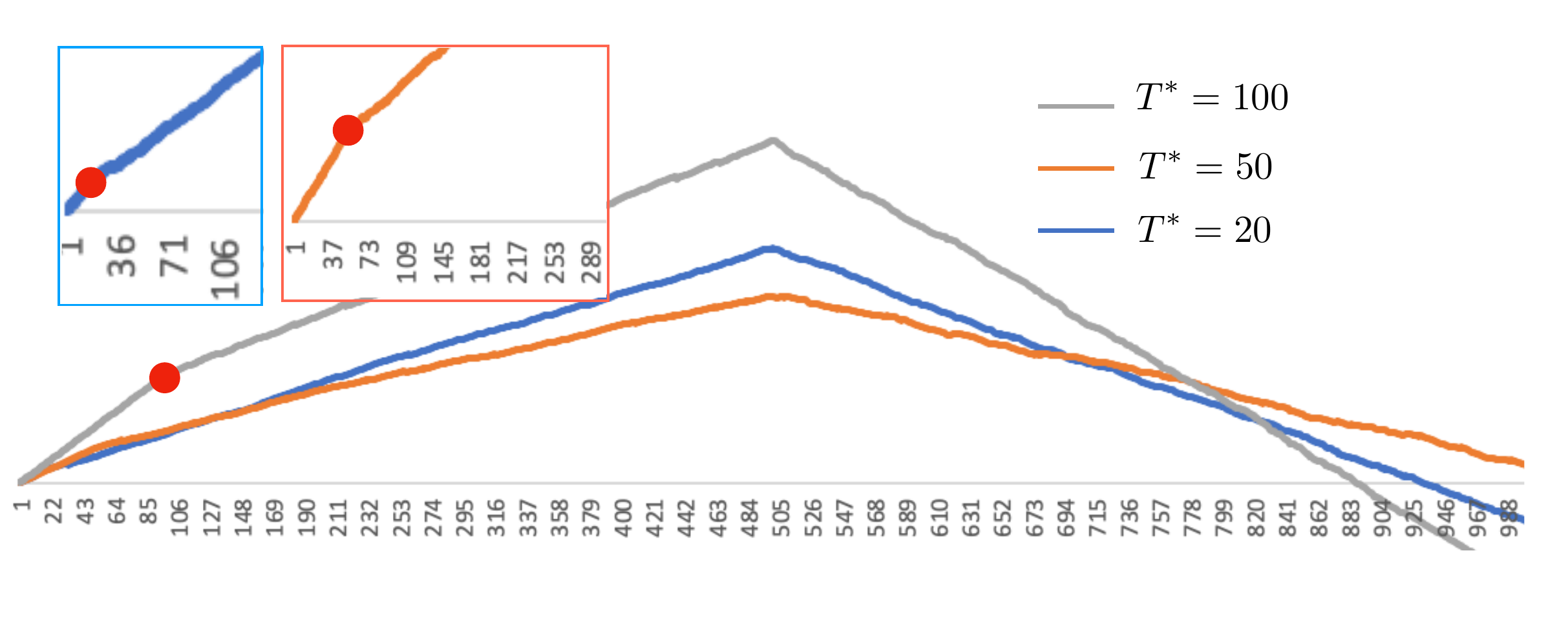}
   \caption{Robustness to $\lvert T^{*} - T_{\text{split}}\rvert$ when $T_{\text{split}} = 500$: $P_1$ and $P_2$  are multivariate Gaussian distributions whose mean vectors $\vec{\mu}_{1}$ and $\vec{\mu}_{2}$ are sampled from $\text{Unif.}[-1,1]$ and $\text{Unif.}[-2,2]$, respectively, with the covariance matrices equal to the identity matrix.}
   \label{fig: Sensitivity DRE-CUSUM-a}
\end{subfigure}

\begin{subfigure}[b]{0.46\textwidth}
   \includegraphics[width=1\linewidth]{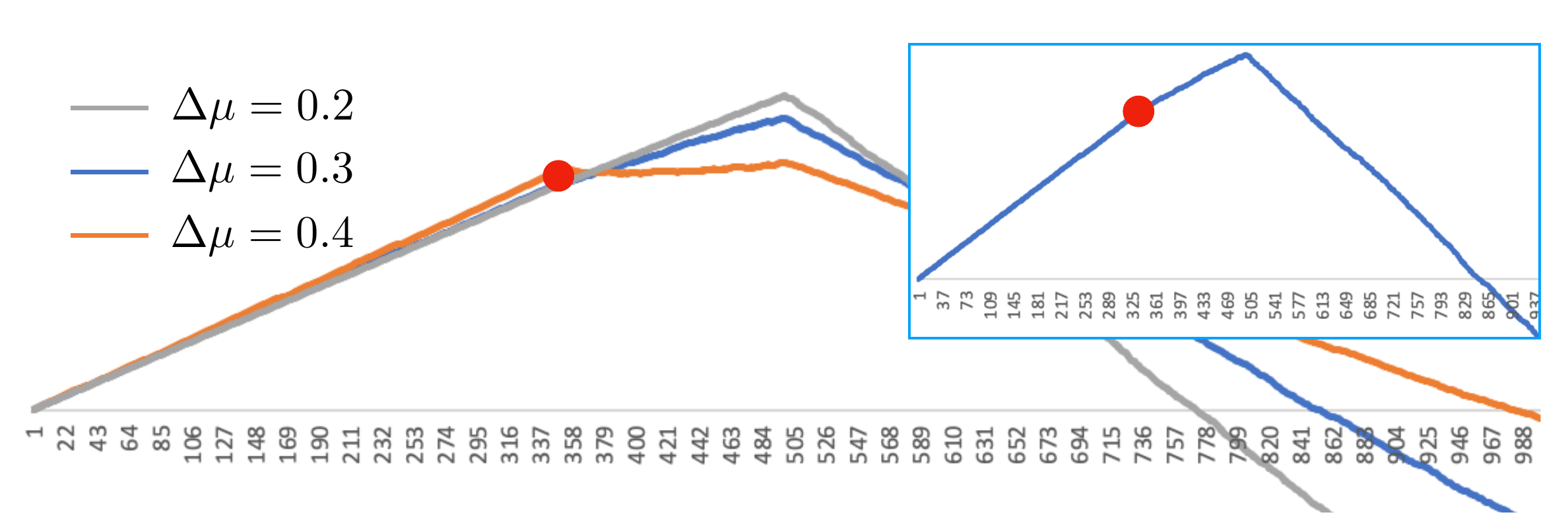}
   \caption{{Robustness to change in distributions: $T_{\text{split}} = 500$ and $T^{*} = 350$. $P_1$ and $P_2$ are multivariate Gaussian distributions with identity covariance matrix, and mean vectors  $\vec{\mu}_{1}$ is sampled from $\text{Unif.}[-1,1]$, and we plot $S_{\text{DR}}^{T_{\text{split}}}(t)$ for different values of $\Delta \vec{\mu}  = \lvert \vec{\mu}_{1} - \vec{\mu}_{2} \rvert $}} 
   \label{fig: Sensitivity DRE-CUSUM-b}
\end{subfigure}
\caption{Robustness of the DRE-CUSUM algorithm}
\vspace{-10pt}
\end{figure}

\noindent {\underline{\textbf{Robustness of DRE-CUSUM}}} To demonstrate the robustness of the DRE-CUSUM to distance $\lvert T^{*} - T_{\text{split}} \rvert$, we consider a $10$-dimensional time-series data with $1000$ samples whose pre-and post-change distributions are sampled from multivariate Gaussian distributions with mean shift at time $T^{*}$ as described in Fig. \ref{fig: Sensitivity DRE-CUSUM-a}. 
We set $T_{\text{split}} = 500$, the change point in the time-series data $T^{*}$ is varied (say $20,50,100$), thereby, varying the number of points in the time-series sampled from distributions $P_1$ and $P_2$.
\textit{From Fig.\ref{fig: Sensitivity DRE-CUSUM-a}, we infer that DRE-CUSUM statistic $S^{T_{\text{split}}}_{\text{DRE}}(t)$ changes slope at $T^{*}$ irrespective of $\lvert T^{*} - T_{\text{split}}\rvert$.}

\noindent For checking the robustness of DRE-CUSUM to distance between $P_1$ and $P_2$, we consider $10-$dimensional time-series data, with a mean shift at time-instance $T^{*} = 350$.
$P_1$ and $P_2$ are  multivariate Gaussian distributions with same covariance matrix. We set mean $\vec{\mu}_1$ corresponding to $P_1$ as shown in Fig. \ref{fig: Sensitivity DRE-CUSUM-b}, and vary the difference $\Delta \vec{\mu} = \lvert \vec{\mu}_{2}-\vec{\mu}_{1} \rvert$.
\textit{From  Fig. \ref{fig: Sensitivity DRE-CUSUM-b}, it is clear that slope of DRE-CUSUM statistic $S^{T_{\text{split}}}_{\text{DRE}}$ changes at $T^{*}=350$ for relatively small $\Delta \vec{\mu}$.} 
We next compare the DRE-CUSUM approach with other unsupervised change detection approaches, particularly Bayesian change detection and its variants \cite{knoblauch2018doubly,fearnhead2006exact}

\begin{table}[!h]
\centering
\setlength{\tabcolsep}{0.5em}
\begin{tabular}{ c c c  } 
 \hline
Methodology & FAR & MDR \\
 \hline
 DRE-CUSUM (DNN, KLIEP)  & \cellcolor[HTML]{D3D3D3}$0\%$ & \cellcolor[HTML]{D3D3D3}$0\%$ \\ 
DRE-CUSUM (DNN, LSIF) & $0\%$ & $14.3\%$ \\ 
DRE-CUSUM (Kernel, LSIF) & $0.0005\%$ & $14.3\%$\\ 
 Online BCD  & $\sim 30\%$ & $\sim 0\%$\\
 Robust Online BCD & $0.04\%$ & $42\%$ \\  
 \hline
\end{tabular}
\captionof{table}{Comparison of online DRE-CUSUM with Online BCD \cite{adams2007bayesian}, and Robust Online BCD  \cite{knoblauch2018doubly}). 
\label{tab:performance online cd}}
\end{table}

\noindent {\underline{\textbf{Comparison with other approaches}}} We consider a 50-dimensional time-series data with 2000 samples generated from multivariate Gaussian distribution with same covariance matrix, such that it undergoes mean changes at $10$ intervals.
The mean vectors of the Gaussian distribution in different segments are sampled from uniform distributions (more details on the dataset generation have been included in the appendix).
The results of DRE-CUSUM (online variant) along with other approaches have been tabulated in Table \ref{tab:performance online cd}, from which we infer that \textit{DRE-CUSUM (for KLIEP objective) outperforms Bayesian approach.}


\begin{table*}[!t]
\centering 
\begin{tabular}{l|c|c|c|c} 
\hline
\, & \multicolumn{2}{c|}{\textbf{HASC}} & \multicolumn{2}{|c}{\textbf{USC}} \\ 
\hline 
Methodology & FAR & MDR & FAR & MDR \\
\hline 
Pelt \cite{wambui2015power} & $0\%$ & $21.42\%$ & \cellcolor[HTML]{D3D3D3}$0.0010\%$ & $74.28\%$\\\
\hspace{-2pt}Dynamic Programming \cite{truong2020selective} & $0.033\%$ & $21.42\%$ & $0.018\%$ & $37.14\%$ \\
Binary segmentation \cite{fryzlewicz2014wild} & $0.033\%$ & $26.67\%$ & $0.024\%$ & $60\%$ \\
\textbf{DRE-CUSUM (Kernel,LSIF)} & \cellcolor[HTML]{D3D3D3}$0\%$ &  \cellcolor[HTML]{D3D3D3}$0\%$ &  $0.0075\%$ &  \cellcolor[HTML]{D3D3D3}$14.2\%$ \\
\hline
\end{tabular}
\caption{Performance comparison of online DRE-CUSUM with Pelt, dynamic programming, and binary segmentation methods on HASC-\cite{ichino2016hasc} and USC- datasets \cite{zhang2012usc,deldari2021time}.} 
\label{table: HASC-USC-dataset} 
\vspace{-10pt}
\end{table*}

\subsection{Experiments on real-world datasets}
In this section, we compare performance of DRE-CUSUM with other unsupervised approaches on two real-world datasets. 
Furthermore, we present the results of the DRE-CUSUM for event detection tasks on video dataset. 
\newline \noindent {\underline{\textbf{Performance comparison on real-world datasets}}} In this section, we perform evaluation against dynamic programming \cite{truong2020selective}, Linearized penalty segmentation (Pelt) \cite{wambui2015power}, and  Binary segmentation (BinSeg) \cite{fryzlewicz2014wild} techniques on the following \textbf{\textit{labeled}} datasets: (i) HASC dataset \cite{ichino2016hasc}, (ii) USC-HAD dataset \cite{zhang2012usc}, which we described next.

For change detection using HASC dataset, we consider the time-series data with $11857$ samples with $11$ change points (for example, activity shift from walking to jogging).
Each datapoint in time-series is $3-$dimensional corresponding to the recordings of the accelerometer along $x,y,z$ axis. 
For change detection using USC dataset, we consider a time-series of $93635$ samples with $35$ change points \cite{zhang2012usc,deldari2021time}. 
Each sample in the time-series is the accelerometer reading along $x-$axis that corresponds to the state at that instance (for example, sitting).
\textit{As seen from Table \ref{table: HASC-USC-dataset}, DRE-CUSUM outperforms its counterparts, particularly Pelt and dynamic programming methods, both of which are known to be accurate in a low-dimensional setting \cite{truong2020selective}. Furthermore, Pelt always has smaller FAR compared to other approaches.}

\begin{figure}[!h]
\centering
\begin{subfigure}[b]{0.46\textwidth}
   \includegraphics[width=1\linewidth]{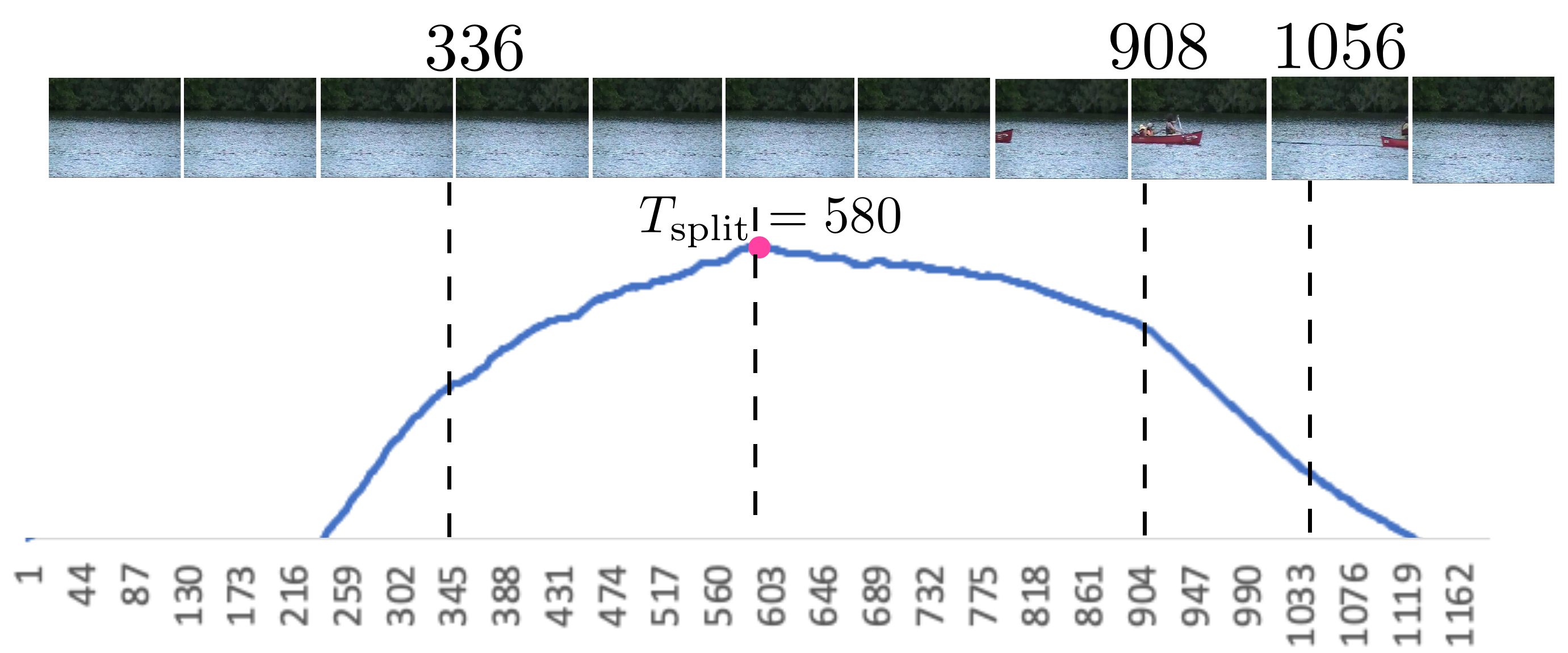}
   \caption{DRE-CUSUM algorithm on Canoe dataset to detect entry and exit of a boat.\label{fig: DRE-CUSUM-Canoe}}
\end{subfigure}

\begin{subfigure}[b]{0.46\textwidth}
   \includegraphics[width=1\linewidth]{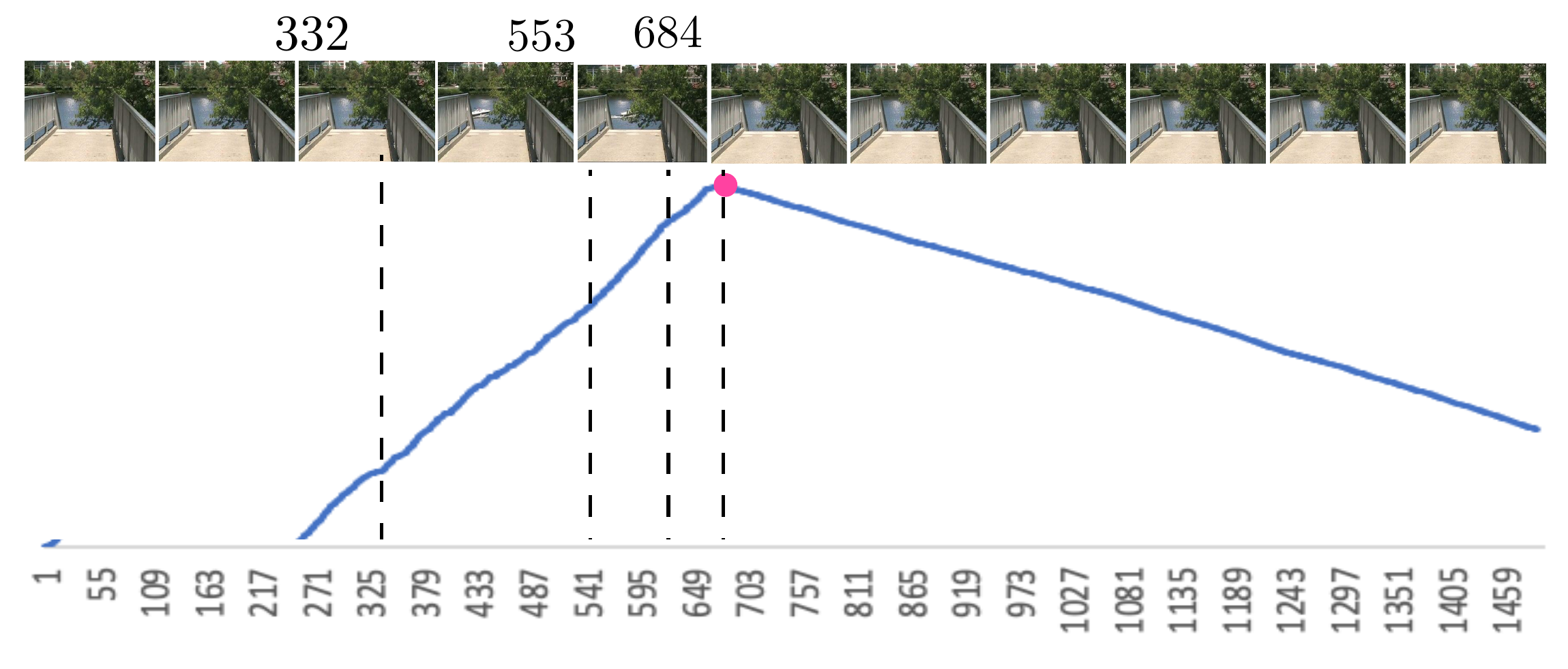}
   \caption{DRE-CUSUM algorithm on Overpass dataset for detecting entry and exit of both person/boat. \label{fig:  DRE-CUSUM-Overpass}} 
\end{subfigure}
\caption{Video event detection using DRE-CUSUM.}
\vspace{-4pt}
\end{figure}

\noindent {\underline{\textbf{Video event detection using DRE-CUSUM}}}
To highlight the main advantage offered by the DRE-CUSUM approach, we consider 2012-Dataset, an \textit{\textbf{unlabeled}} high dimensional video dataset \cite{goyette2012changedetection}.
We applied the DRE-CUSUM on two video/image sequences: canoe and overpass, with the objective of finding instances that demarcate the start or end of an event (for example, entry/exit of a boat). 
Fig. \ref{fig: DRE-CUSUM-Canoe} and Fig. \ref{fig: DRE-CUSUM-Overpass} correspond to DRE-CUSUM statistics for video frames from canoe and overpass dataset, respectively, and the details of the experimental setup are as described next. 
\\ (a) Canoe dataset: The time-series in Fig. \ref{fig: DRE-CUSUM-Canoe} has $1189$ video frames. We set $T_\text{split} = 580$.
Frames $908$, and $1056$ marks the entry, and  the exit of the boat, respectively. At the corresponding instances, we observe slope changes in DRE-CUSUM statistic.
On visual inspection, we note that there are no significant changes at frame $336$ (the slope change at t = $336$ in DRE-CUSUM is observed for different values of $T_{\text{split}}$). We declare the slope change at $336$ as a false alarm.    
\\ (b) Overpass dataset: In Fig. \ref{fig: DRE-CUSUM-Overpass}, the time-series has $1500$ samples, wherein we set $T_\text{split} = 700$.
Slope changes present in DRE-CUSUM statistic around frames $553$ and $684$ corresponds to the object entry and exit frames, respectively. 
However, the slope change around the frame $332$ is a false alarm.

Additional experimental results on video sequences in 2012-Dataset are provided in Appendix \ref{sec: experiments-appendix}. 
\textit{We can extrapolate from these experimental results that: one can in principle use DRE-CUSUM to determine the change point estimates across any time-series data (both high-and low-dimensional). The slope changes can then be used for interpretation.} 

\section{Discussion and Future Work}
\label{sec: discussion}
In this paper, we proposed DRE-CUSUM, a novel approach for unsupervised change detection, and showed its broad applicability on a wide range of applications backed by theoretical guarantees and experimental results. The salient aspect of DRE-CUSUM is that it does not require any knowledge/specification of the underlying distributions, nor an estimate of the number of underlying change points, and is universally applicable for high-dimensional data. 
To the best of our knowledge, our work is the first to provide theoretical justification and accuracy guarantees for the use of density ratio based unsupervised change detection. 
There are several possible directions for future work and we list some of them below:
\newline \noindent a) Obtaining accuracy guarantees for DRE-CUSUM when there are multiple change points is an immediate interesting research direction.   
\newline \noindent b) The accuracy guarantees for online adaptation of the DRE-CUSUM can be derived to understand the theoretical trade-off between between fixed vs. adaptive window size for a given time-series data. 
\newline \noindent c) In real-world applications, we use density ratio estimators (DREs) to implement DRE-CUSUM. Studying the impact of sample complexity (i.e. number of samples required for a good estimation of the density ratio) of different DRE models on change point detection accuracy is also an important direction. 

\bibliographystyle{plain}
\bibliography{refs.bib}

\appendices

\section{Proof of Proposition $1$}
\label{sec:Theorem-1-Proof}
\begin{proof}
For the case when $T_\text{{split}}\leq T^{*}$, we have the following two distributions before and after $T_\text{{split}}$, respectively: 
\begin{align}
P_{\text{left}} = P_1, \hspace{1.5cm} P_{\text{right}} = \alpha^{T_{\text{split}}} P_1 + (1 - \alpha^{T_{\text{split}}}) P_2,
\end{align}
where, $\alpha^{T_{\text{split}}} = \frac{T^{*}-T_{\text{split}}}{n-T_{\text{split}}}$. 
The expected values of the log-likelihood ratio $P_{\text{left}}(.)/P_{\text{right}}(.)$ before $T^{*}$ (i.e., $\forall t < T^{*}$) is,
\begin{align}
 &\mathbb{E}_{x_t}\left[\log \left(\frac{P_{\text{left}}(x_t)}{P_{\text{right}}(x_t)}\right)\right]  \nonumber \\&=    \mathbb{E}_{x_t \sim P_1} \left[ \log \left(\frac{P_1(x_t)}{\alpha^{T_{\text{split}}} P_1(x_t) + (1-\alpha^{T_{\text{split}}}) P_2(x_t)} \right) \right] 
\nonumber \\&= \int_{x} P_1(x)  \log \left( \frac{P_1(x)}{\alpha^{T_{\text{split}}} P_1(x) + (1-\alpha^{T_{\text{split}}}) P_2(x)}\right) dx
\nonumber \\ & \stackrel{(a)}{=} KL(P_1||P_{\text{right}}) 
\label{eq: equation-1-supp}
\end{align}
where, (a) follows from the definition of KL-divergence. 
For simplicity of notation, let us  define $\hspace{0.3cm} \alpha_1 \triangleq \frac{n-T^{*}}{n-T_{\text{split}}}$, therefore, $\alpha^{T_{\text{split}}} = 1 - \alpha_1$, and from definition of the parametric mixture distribution, we can write $P_{\text{right}} = P(1 - \alpha_1)$. 
Substituting in \eqref{eq: equation-1-supp} we get,
\begin{align}
 \mathbb{E}_{x_t}\left[\log \left(\frac{P_{\text{left}}(x_t)}{P_{\text{right}}(x_t)}\right)\right] = KL(P_1||P(1 - \alpha_1)) \geq 0
\end{align}
Similarly, we obtain the expected values of the log-likelihood ratio $P_{\text{left}}(.)/P_{\text{right}}(.)$ for any $t\geq T^{*}$ as follows:
\begin{align}
  \mathbb{E}_{x_t}&\left[\log \left(\frac{P_{\text{left}}(x_t)}{P_{\text{right}}(x_t)}\right)\right] 
  \nonumber \\=& \mathbb{E}_{x_t \sim P_2} \bigg[\log \left( \frac{P_1(x_t)}{\alpha^{T_{\text{split}}} P_1(x_t)+(1-\alpha^{T_{\text{split}}}) P_2(x_t)} \right) \bigg] \nonumber \\
=& -\mathbb{E}_{x_t \sim P_2} \bigg[\log \left(  \frac{\alpha^{T_{\text{split}}} P_1(x_t)+(1-\alpha^{T_{\text{split}}}) P_2(x_t)}{P_1(x_t)}\right)\bigg]\nonumber \\
=& -\int_{x}P_2(x) \log \left( \frac{P_{\text{right}}(x)}{P_1(x)}\right)dx\nonumber \\
\stackrel{(a)}{=}& -\frac{1}{1-\alpha^{T_{\text{split}}}} \bigg[\int_{x} P_{\text{right}}(x)   \log \left( \frac{P_{\text{right}}(x)}{P_1(x)}\right) dx \nonumber\\ &\hspace{2cm}+ \int_{x} \alpha^{T_{\text{split}}} P_1(x) \log \left(\frac{P_1(x)}{P_{\text{right}}(x)}\right) dx  \bigg]
\nonumber\\
\stackrel{(b)}{=}&-\left[\frac{1}{1-\alpha^{T_{\text{split}}}} KL(P_\text{right} || {P_1}) + \frac{\alpha^{T_{\text{split}}} }{1-\alpha^{T_{\text{split}}}}  KL({P_1}||P_\text{right})\right] 
\nonumber\\
\stackrel{(c)}{=}&-\left[\frac{1}{\alpha_1} KL((1- \alpha_1)P_1 +  \alpha_1 P_2 || P_1)\right. \nonumber\\ &\hspace{1cm}\left.+ \frac{1-\alpha_1}{ \alpha_1}  KL(P_1||(1- \alpha_1)P_1 +  \alpha_1 P_2)\right] 
\nonumber\\
\stackrel{(d)}{=}&-f_{1}\left(\alpha_1,(1-\alpha_1)\right) \leq 0 
\label{eq: equation-4}
\end{align}
where, (a) and (c) follows from substituting $P_2(x_t) = \left(P_{\text{right}}(x_t) - \alpha^{T_{\text{split}}}P_1(x_t)\right)/(1-\alpha^{T_{\text{split}}})$ and $\alpha_1 = 1-\alpha^{T_{\text{split}}}$, respectively. 
(b) and (d) follow from definition of KL-divergence and definition 1 in the paper, respectively.  
\\
\noindent For the case when $T_\text{{split}}\geq T^{*}$, we have the following two distributions before and after $T_\text{{split}}$ respectively: 
\begin{align}
   P_{\text{left}} = \alpha^{T_{\text{split}}} P_1 + (1 - \alpha^{T_{\text{split}}}) P_2, \hspace{1.5cm} P_{\text{right}} = P_2,
\end{align}
where, $\alpha^{T_{\text{split}}} = \frac{T^{*}}{T_{\text{split}}}$.
Expected value of the log-likelihood ratio $P_{\text{left}}(.)/P_{\text{right}}(.)$ before the change point $T^{*}$ (i.e., $\forall t < T^{*}$) is:
\begin{align}
& \mathbb{E}_{x_t} \left[ \log \left(\frac{P_{\text{left}}(x_t)}{P_{\text{right}}(x_t)}\right)\right]
\nonumber \\ &=  \mathbb{E}_{x_t \sim P_1} \left[ \log \frac{\alpha^{T_{\text{split}}} P_1(x_t) + (1-\alpha^{T_{\text{split}}}) P_2(x_t)}{P_2(x_t)} \right] \nonumber\\
&= \int_{x} P_1(x)  \log \frac{P_{\text{left}}(x)}{P_2(x)} dx \nonumber\\ 
& \stackrel{(a)}{=}\frac{1}{\alpha^{T_{\text{split}}}} \bigg[\int_{x} 
P_{\text{left}}(x)   \log \frac{P_{\text{left}}(x)}{P_2(x)} dx \nonumber\\& \hspace{1.5cm}- \int_{x} (1-\alpha^{T_{\text{split}}}) P_2(x) \log \frac{P_{\text{left}}(x)}{P_2(x)} dx  \bigg] \nonumber\\
& \stackrel{(b)}{=}\frac{1}{\alpha^{T_{\text{split}}}} KL(P_{\text{left}} || {P_2}) + \frac{1-\alpha^{T_{\text{split}}}}{\alpha^{T_{\text{split}}}}  KL({P_2}||P_{\text{left}}) 
\label{eq: summary-theorem-1-case-A-i}
\end{align}
where, (a) follows from the fact that $P_1(x_t) = \left(P_{\text{left}}(x_t)-(1-\alpha^{T_{\text{split}}})P_{2}(x_t)\right)/\alpha^{T_{\text{split}}}$, (b) follows from the definition of KL-divergence.
For simplicity of notation, let us define $\alpha_2 \triangleq \alpha^{T_{\text{split}}}$, and
subsequently $\forall t \geq T^{*}$, \eqref{eq: summary-theorem-1-case-A-i} becomes 
\begin{align}
\mathbb{E}_{x_t}\bigg[\log  & \left(\frac{P_{\text{left}}(x_t)}{P_{\text{right}}(x_t)}\right)\bigg] \nonumber \\&=\frac{1}{ \alpha_2} KL(P_{\text{left}} || {P_2}) + \frac{1-\alpha_2}{\alpha_2}  KL({P_2}||P_{\text{left}}) \nonumber \\ 
&= f_2(\alpha_2,\alpha_2) \geq 0.
\end{align}
The expected values of the log-likelihood ratio $P_{\text{left}}(.)/P_{\text{right}}(.)$ for all points after $T^{*}$ (i.e., $\forall t \geq T^{*}$) can be computed as
\begin{align}
\mathbb{E}_{x_t} & \left[\log\left(\frac{P_{\text{left}}(x_t)}{P_{\text{right}}(x_t)}\right)\right] \nonumber \\ &= \mathbb{E}_{x_t \sim P_2} \bigg[\log \frac{\alpha^{T_{\text{split}}} P_1(x_t)+(1-\alpha^{T_{\text{split}}}) P_2(x_t)}{P_2(x_t)}\bigg]  \nonumber \\ &= 
-KL(P_2||P_{\text{left}}) = -KL(P_2||P(\alpha_2))\leq 0.
\end{align}
\textit{As a consequence of the above results, we infer that for any choice of $T_{\text{split}}$} 
\begin{align}
    {\mathbb{E}_{x_t}}\log \left(\frac{P_{\text{left}}(x_t)}{P_{\text{right}}(x_t)}\right) =
    \begin{cases}
        \geq 0, & \text{for } t < T^{*} \\
        \leq 0, & \text{for } t \geq T^{*}
    \end{cases}
    \label{eq: inference-theorem}
\end{align}
\textit{From the above result, Corollary $1$ stated in the paper follows immediately, since the statistic $S^{T_{\text{split}}}_{\text{DR}}(t)$ is a linear function in $t$ with a non-negative slope before $T^{*}$, and conversely, a linear function in $t$ with a non-positive slope for all points after $T^{*}$.}
\end{proof}

\section{Proof of Theorem 1}
\label{sec:Theorem-2-Proof}
\begin{figure}[!h]
\centering
\begin{subfigure}[b]{0.45\textwidth}
   \includegraphics[width=0.85\linewidth]{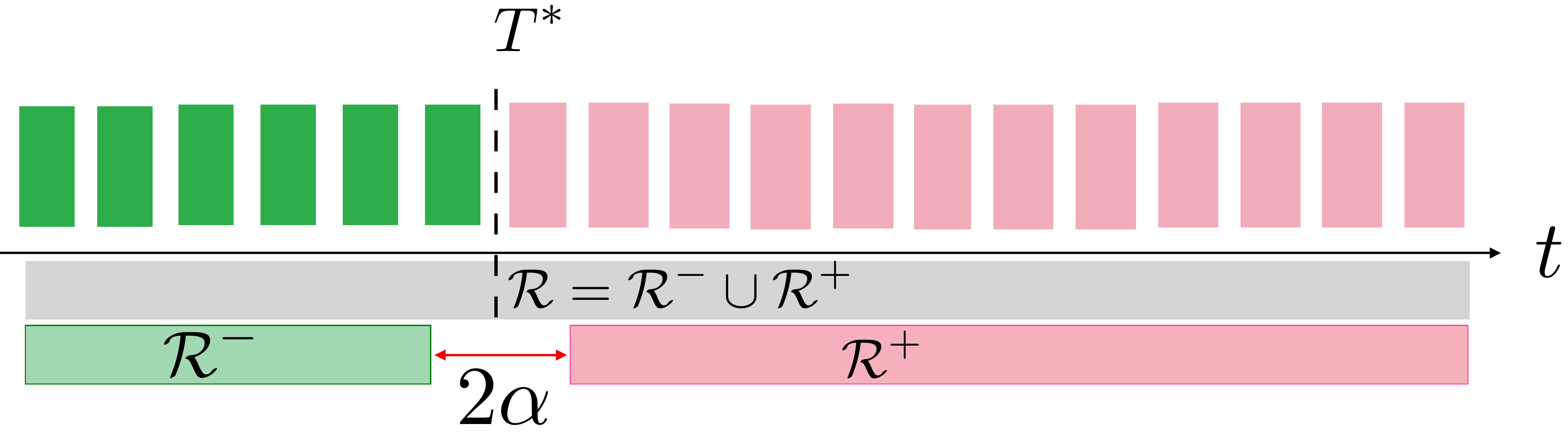}
	\caption{Regions $\mathcal{R^{-}}$ and $\mathcal{R^{+}}$, such that $\mathcal{R} = \mathcal{R^{-}} \cup \mathcal{R^{+}}$. \label{fig: Region-1-2-Thm-2}}
\end{subfigure}
\begin{subfigure}[b]{0.45\textwidth}
   \includegraphics[width=1\linewidth]{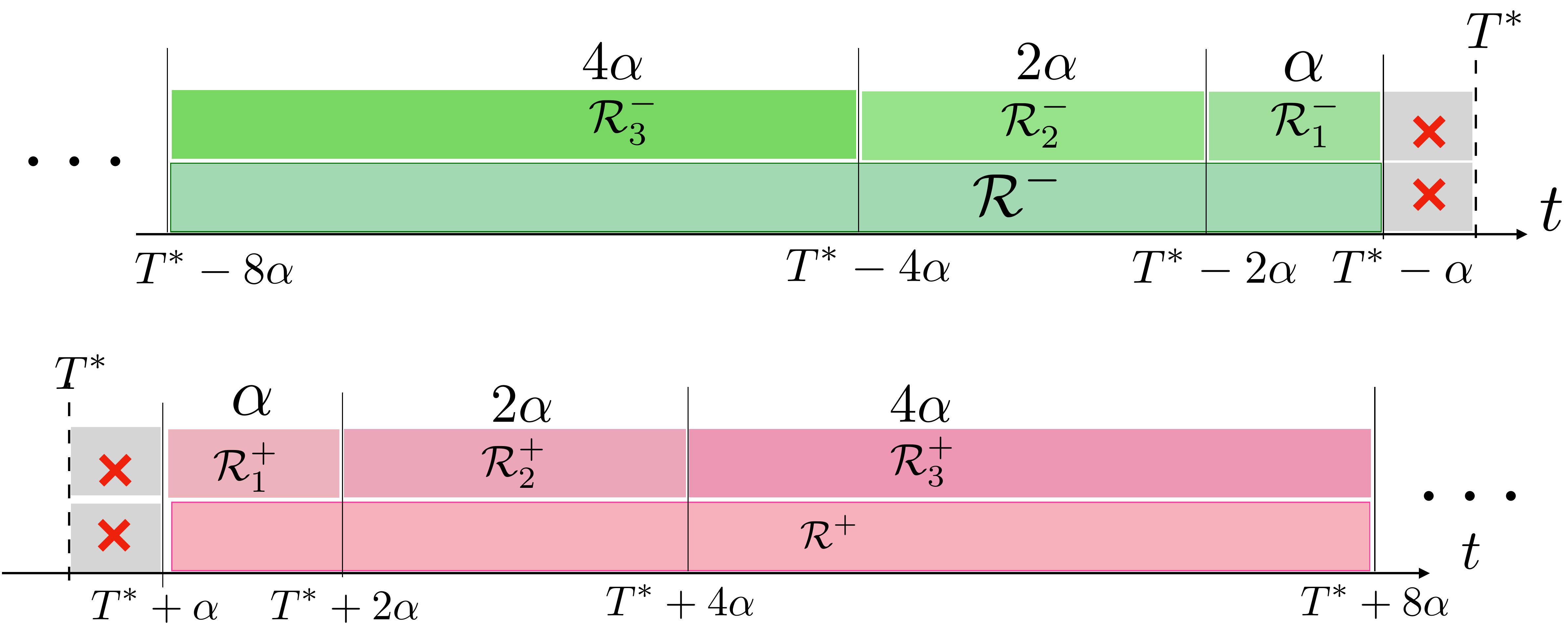}
  \caption{Sub-intervals of increasing lengths in regions $\mathcal{R}^{-}$ and $\mathcal{R}^{+}$. \label{fig: Intervals_r-1-2-Thm-2}}
\end{subfigure}
\caption{Region $\mathcal{R}$ split into to sub-regions $\mathcal{R}^{-} = [1,T^{*}-\alpha)$ and $\mathcal{R}^{+} = (T^{*}+\alpha,n]$ in Fig. 1(a), which is further segmented to smaller regions as shown in Fig. 1(b).}
\vspace{-10pt}
\end{figure}

To prove that the change estimate $\hat{T}_{\text{DR-CUSUM}}$ is $\left(\alpha,\beta\right)$- accurate, we use similar reasoning to that used in \cite{cummings2018differentially} to show that the maximum likelihood estimate $\hat{T}_{\text{ML}}$ is $\left(\alpha,\beta\right)$- accurate.
However, the maximum likelihood approach assumes that $P_1{(.)}/P_2{(.)}$ can be readily computed. 
The key distinction of our proof is that  we show the change point estimate using DR-CUSUM $\hat{T}_{\text{DR-CUSUM}}$  is $\left(\alpha,\beta\right)$- accurate even when $P_{\text{left}}{(.)}/P_{\text{right}}{(.)}$ can be computed in the unsupervised setting. 
To prove this result, we consider two separate cases: (a)  $T_{\text{split}} \leq T^{*}$ and (b) $T_{\text{split}} \geq T^{*}$. For case (a), when $T_{\text{split}} \leq T^{*}$, we have
\begin{align}
  P_{\text{left}} = P_1, \hspace{1.3cm} P_{\text{right}} = \alpha^{T_{\text{split}}} P_1 + \left(1-\alpha^{T_{\text{split}}}\right) P_2.  
    \label{eq: theorem-2-eqn-0}
\end{align}
The density-ratio CUSUM statistic $S^{T_{\text{split}}}_{\text{DR}}(t)$ is defined $\forall t \in [1,n]$ as:
\begin{align}
    S^{T_{\text{split}}}_{\text{DR}}(t) \overset{\Delta}{=} \sum_{j=1}^{t} \left(\log \left(\frac{P_{\text{left}}(x_j)}{P_{\text{right}}(x_j)}\right)\right) 
    \label{eq: theorem-2-eqn-1}
\end{align}
For a given $\alpha$, let us define the region $\mathcal{R}$ as:
\begin{align}
    \mathcal{R} = [n]\backslash [T^{*}-\alpha,T^{*}+\alpha]
    \label{eq: theorem-2-eqn-2}
\end{align}
which essentially consists of all time instances which are at least a distance $\alpha$ from the true change point $T^{*}$. 
We first note the following inequality:
\begin{align}
    P[|\hat{T}_{\text{DR-CUSUM}} & - T^{*}|  > \alpha] \nonumber \\ & \leq  P(\underset{t \in \mathcal{R}}{{\max}} \text{ } S^{T_{\text{split}}}_{\text{DR}}(t)-S^{T_{\text{split}}}_{\text{DR}}(T^{*}) > 0) \label{eq: theorem-2-eqn-3}
\end{align}
which is a direct consequence of the definition of $\mathcal{R}$ and the DRE-CUSUM estimator. From now on, our goal will be to upper bound the probability $P(\underset{t \in \mathcal{R}}{{\max}} \text{ } S^{T_{\text{split}}}_{\text{DR}}(t)-S^{T_{\text{split}}}_{\text{DR}}(T^{*}) > 0)$. To this end, we consider regions  $\mathcal{R}^{-} = \left[1,T^{*}-\alpha \right)$ and $\mathcal{R}^{+} = \left(T^{*}+\alpha, n\right]$ as shown in Fig. \ref{fig: Region-1-2-Thm-2}, such that $\mathcal{R} = \mathcal{R}^{+} \cup \mathcal{R}^{-}$.
Thereafter, on applying union-bound to the r.h.s. of \eqref{eq: theorem-2-eqn-3} over regions $\mathcal{R}^{-}$ and $\mathcal{R}^{+}$, we have
\begin{align}
     P(\underset{t \in \mathcal{R}}{{\max}} \text{ } S^{T_{\text{split}}}_{\text{DR}}(t) &- S^{T_{\text{split}}}_{\text{DR}}(T^{*}) > 0)  \nonumber \\  \leq &  P(\underset{t \in \mathcal{R}^{-}}{{\max}} \text{ } S^{T_{\text{split}}}_{\text{DR}}(t)- S^{T_{\text{split}}}_{\text{DR}}(T^{*}) > 0) \nonumber \\ & + P(\underset{t \in \mathcal{R}^{+}}{{\max}} \text{ } S^{T_{\text{split}}}_{\text{DR}}(t)-S^{T_{\text{split}}}_{\text{DR}}(T^{*}) > 0)
    \label{eq: theorem-2-eqn-4}
\end{align}
We have the following two cases to compute $S^{T_{\text{split}}}_{\text{DR}}(t)- S^{T_{\text{split}}}_{\text{DR}}(T^{*})$:
\begin{align}
    S^{T_{\text{split}}}_{\text{DR}}&(t) - S^{T_{\text{split}}}_{\text{DR}}(T^{*})  \nonumber \\ =&
    \begin{cases}
    -\sum_{i=t}^{T^{*}-1} \log \left[{P_{\text{left}}(x_i)}/{P_{\text{right}}(x_i)}\right], & t \in \mathcal{R}^{-} \\
    -\sum_{i=T^{*}}^{t-1} \log \left[{P_{\text{right}}(x_i)}/{P_{\text{left}}(x_i)}\right], & t \in \mathcal{R}^{+}
    \end{cases}
    \label{eq: theorem-2-eqn-5}
\end{align}
\newline \noindent We further segment regions $\mathcal{R}^{-}$ and $\mathcal{R}^{+}$, such that the interval lengths double in length as we move away from change point $T^{*}$ as shown in Fig. \ref{fig: Intervals_r-1-2-Thm-2}.    
\textit{Assuming finite samples in the time-series data, the total number intervals in $\mathcal{R}^{-}$, and $\mathcal{R}^{+}$ are $\log_2 \left(\frac{T^{*}}{\alpha}\right)$, and $\log_2 \left(\frac{n-T^{*}}{\alpha}\right)$, respectively}.
Therefore, we can simplify \eqref{eq: theorem-2-eqn-4} further as:
\begin{align}
   & P( \underset{t \in \mathcal{R}}{{\max}} \text{ } S^{T_{\text{split}}}_{\text{DR}}(t)-S^{T_{\text{split}}}_{\text{DR}}(T^{*}) > 0) \nonumber \\  &\stackrel{(a)}{\leq}  
        P\left(\underset{t \in \mathcal{R}^{-}}{{\max}}  -\sum_{i=t}^{T^{*}-1} \log \frac{P_{\text{left}}(x_i)}{P_{\text{right}}(x_i)} > 0\right) \nonumber \\ & \hspace{0.4cm} +  P\left(\underset{t \in \mathcal{R}^{+}}{{\max}} -\sum_{i=T^{*}}^{t-1} \log \frac{P_{\text{right}}(x_i)}{P_{\text{left}}(x_i)}> 0 \right) \nonumber \\ 
      &\stackrel{(b)}{\leq} \underbrace{\sum_{i=1}^{\log_2 \left(\frac{T^{*}}{\alpha}\right)}  P\left(\underset{t \in \mathcal{R}_{i}^{-}}{{\max}}  -\sum_{i=t}^{T^{*}-1} \log \left(\frac{P_{\text{left}}(x_i)}{P_{\text{right}}(x_i)}\right) > 0\right)}_{S_1} \nonumber \\ & \hspace{0.5cm} + \underbrace{\sum_{i=1}^{\log_2 \left(\frac{n-T^{*}}{\alpha}\right)} P\left(\underset{t \in \mathcal{R}_{i}^{+}}{{\max}} -\sum_{i=T^{*}}^{t-1} \log \left(\frac{P_{\text{right}}(x_i)}{P_{\text{left}}(x_i)}\right)> 0\right)}_{S_2}
    \label{eq: theorem-2-eqn-6}
\end{align}

\noindent where, (a) follows from substituting \eqref{eq: theorem-2-eqn-5} in \eqref{eq: theorem-2-eqn-4}, and (b) follows from applying union bound over segments in regions $\mathcal{R}^{-}$ and $\mathcal{R}^{+}$ depicted in Fig. \ref{fig: Intervals_r-1-2-Thm-2}.  
We individually upper bound the terms $S_1$, and $S_2$ in \eqref{eq: theorem-2-eqn-6}. 
The term $S_1$ in \eqref{eq: theorem-2-eqn-6}, when $T_{\text{split}} \leq T^{*}$ can be simplified as follows:
\begin{align}
&\sum_{i=1}^{\log_2 \left(\frac{T^{*}}{\alpha}\right)}    P \left(\underset{t \in \mathcal{R}_{i}^{-}}{{\max}} -\sum_{i=t}^{T^{*}-1} \log \left(\frac{P_{\text{left}}(x_i)}{P_{\text{right}}(x_i)}\right) > 0\right)
\nonumber \\ & \stackrel{(a)}{\leq} \sum_{i=1}^{\log_2 \left(\frac{T^{*}}{\alpha}\right)}  P \bigg(\underset{t \in \mathcal{R}_{i}^{-}}{{\max}} -\sum_{i=t}^{T^{*}-1} \log \left(\frac{P_{\text{left}}(x_i)}{P_{\text{right}}(x_i)}\right) \nonumber \\ & \hspace{3.2cm} + \left(T^{*}-t\right) KL(P_1||P(1-\alpha_1)) \nonumber \\ & \hspace{3.2cm} -  \left(T^{*}-t\right) KL(P_1||P(1-\alpha_1)) > 0 \bigg)
\nonumber \\ & \stackrel{(b)}{\leq} \sum_{i=1}^{\log_2 \frac{T^{*}}{\alpha}}  P(\underset{t \in \mathcal{R}_{i}^{-}}{{\max}} \lvert \sum_{j=t}^{T^{*}-1} u(x_j)  \rvert > 2^{i-1}\alpha KL(P_1||P(1-\alpha_1))) 
\nonumber \\ & \stackrel{(c)}{\leq} \sum_{i=1}^{\log_2 \left(\frac{T^{*}}{\alpha}\right)} \frac{2\exp{\left(-2^{i-2} \alpha \left(KL(P_1||P(1-\alpha_1))\right)^2/A^2\right)}}{1-2\exp{\left(-2^{i-2} \alpha \left(KL(P_1||P(1-\alpha_1))\right)^2/A^2\right)}}
\nonumber \\ & \stackrel{(d)}{\leq} \sum_{i=1}^{\log_2 \left(\frac{T^{*}}{\alpha}\right)} 4\exp{\left(-2^{i-2} \alpha \left(KL(P_1||P(1-\alpha_1))\right)^2/A^2\right)}
\label{eq: theorem-2-eqn-7}
\end{align}
where, (a) follows by adding and subtracting  $\left(T^{*}-t\right)KL(P_1||P(1-\alpha_1)) $, (b) follows from the fact that for any interval $\mathcal{R}^{-}_{i} = \left(T^{*}-2^{i}\alpha,T^{*}-2^{i-1}\alpha\right]$, we have $T^{*}-t \geq 2^{\left(i-1\right)}\alpha$ and from definition $u(x_j) = -\log \left[P_{\text{left}} (x_j)/P_{\text{right}}(x_j)\right] + KL(P_1||P(1-\alpha_1)) $, for any $x_j$ such that $u(.)$ has zero mean, (c) follows directly from application of Corollary 2 in \cite{cummings2018differentially}, which states that: For $S_k = \sum_{i\in[k]} u_i$ for $k \in [m]$, where $u_1,\cdots,u_m$ are i.i.d random variables with mean zero and strictly bounded by interval of length $L$, we have
\begin{align}
\operatorname{Pr}\left[\max _{k \in[m]}\left|S_{k}\right|>\lambda_{1}+\lambda_{2}\right] \leq \frac{2 \exp \left(-2 \lambda_{1}^{2} /\left(m L^{2}\right)\right)}{1-2 \exp \left(-2 \lambda_{2}^{2} /\left(m L^{2}\right)\right)}
\label{eq: corollary-2-cummings}
\end{align}
where, $\lambda_1, \lambda_2 > 0$. 
The concentration inequality in \eqref{eq: corollary-2-cummings} stems by further upper bounding Ottaviani’s inequality using Hoeffding’s inequality under the assumption that $u_j$ can only take values from interval of bounded length $L$.
Furthermore, in step (c) in \eqref{eq: theorem-2-eqn-7}, as the individual terms represent probabilities, we can upper bound the terms:
\begin{align}
& \frac{2\exp{\left(-2^{i-2} \alpha \left(KL(P_1||P(1-\alpha_1))\right)^2/A^2\right)}}{1-2\exp{\left(-2^{i-2} \alpha \left(KL(P_1||P(1-\alpha_1))\right)^2/A^2\right)}} \leq 1 \nonumber \\ \implies &  {\exp{\left(-2^{i-2} \alpha \left(KL(P_1||P(1-\alpha_1))\right)^2/A^2\right)}} < \frac{1}{4}
\label{eq: theorem-2-eqn-8}
\end{align}
Following similar steps, we can bound the term $S_2$ in \eqref{eq: theorem-2-eqn-6} when $T_{\text{split}} \leq T^{*}$ as
\begin{align}
& \sum_{i=1}^{\log_2 \left(\frac{n-T^{*}}{\alpha}\right)}  P\left(\underset{t \in \mathcal{R}_{i}^{+}}{{\max}} -\sum_{j=T^{*}}^{t-1} \log \left(\frac{P_{\text{right}}(x_j)}{P_{\text{left}}(x_j)}\right) > 0\right)
\nonumber \\ & \stackrel{(a)}{\leq}\sum_{i=1}^{\log_2 \left(\frac{n-T^{*}}{\alpha}\right)} P\bigg(\underset{t \in \mathcal{R}_{i}^{+}}{{\max}}  -\sum_{j=T^{*}}^{t-1} \log \left(\frac{P_{\text{right}}(x_j)}{P_{\text{left}}(x_j)}\right) \nonumber \\ & \hspace{3.5cm} + \left(t-T^{*}\right)f_{1}(\alpha_1,1-\alpha_1)  \nonumber \\ & \hspace{3.5cm} -  \left(t-T^{*}\right)f_{1}(\alpha_1,1-\alpha_1)  > 0\bigg) 
\nonumber \\ & \stackrel{(b)}{\leq} \sum_{i=1}^{\log_2 \left(\frac{n-T^{*}}{\alpha}\right)}  P\left(\underset{t \in \mathcal{R}_{i}^{+}}{{\max}} \lvert \sum_{j=T^{*}}^{t-1} u(x_j)  \rvert > 2^{(i-1)}\alpha f_{1}(\alpha_1,1-\alpha_1)\right) 
\nonumber \\ & \stackrel{(c)}{\leq} \sum_{i=1}^{\log_2 \left(\frac{n-T^{*}}{\alpha}\right)} \frac{2\exp{\left(-2^{i-2} \alpha f^{2}_{1}(\alpha_1,1-\alpha_1)/A^2\right)}}{1-2\exp{\left(-2^{i-2} \alpha f^{2}_{1}(\alpha_1,1-\alpha_1)/A^2\right)}}
\nonumber \\ & \stackrel{(d)}{\leq}  \sum_{i=1}^{\log_2 \left(\frac{n-T^{*}}{\alpha}\right)} 4\exp{\left(-2^{i-2} \alpha f^{2}_{1}(\alpha_1,1-\alpha_1)/A^2\right)}
\label{eq: theorem-2-eqn-9}
\end{align}
The steps (a,b,c,d) in \eqref{eq: theorem-2-eqn-9} follow the same logical reasoning to steps (a,b,c) in \eqref{eq: theorem-2-eqn-7}.
Step (e) follows from the same logical reasoning as \eqref{eq: theorem-2-eqn-8}, which is.
\begin{align}
     & \frac{2\exp{\left(-2^{i-2} \alpha f^{2}_{1}(\alpha_1,1-\alpha_1)/A^2\right)}}{1-2\exp{\left(-2^{i-2} \alpha f^{2}_{1}(\alpha_1,1-\alpha_1)/A^2\right)}} \leq 1  \nonumber \\ \implies & {\exp{\left(-2^{i-2} \alpha f^{2}_{1}(\alpha_1,1-\alpha_1)/A^2\right)}} < \frac{1}{4}
\label{eq: theorem-2-eqn-10}
\end{align}
Substituting \eqref{eq: theorem-2-eqn-7} and \eqref{eq: theorem-2-eqn-9} in \eqref{eq: theorem-2-eqn-6} we get,
\begin{align}
& P(\underset{t \in \mathcal{R}}{{\max}}\text{ } S^{T_{\text{split}}}_{\text{DR}}(t)-S^{T_{\text{split}}}_{\text{DR}}(T^{*})  > 0) \nonumber \\ &  \leq \sum_{i=1}^{\frac{T^{*}}{\alpha}} 4\exp{\left(-2^{i-2} \alpha \left(KL(P_1||P(1-\alpha_1))\right)^2/A^2\right)}
   \nonumber \\ & \hspace{1cm} + \sum_{i=1}^{\log_2 \left(\frac{n-T^{*}}{\alpha}\right)} 4\exp{\left(-2^{i-2} \alpha f^{2}_{1}(\alpha_1,1-\alpha_1)/A^2\right)}
\nonumber \\ &
 \stackrel{(a)}{=} \sum_{i=1}^{\frac{T^{*}}{\alpha}}  4\exp{\left(- \alpha \left(KL(P_1||P(1-\alpha_1))\right)^2/2A^2\right)^{2^{i-1}}} 
   \nonumber \\ & \hspace{1cm} + \sum_{i=1}^{\log_2 \left(\frac{n-T^{*}}{\alpha}\right)}  4\exp{\left(- \alpha  f^{2}_{1}(\alpha_1,1-\alpha_1)/2A^2\right)^{2^{i-1}}}
\nonumber \\ &
  \stackrel{(b)}{\leq} \sum_{i=1}^{\frac{T^{*}}{\alpha}}  4\exp{\left(- \alpha \left(KL(P_1||P(1-\alpha_1))\right)^2/2A^2\right)^{i}} 
   \nonumber \\ & \hspace{1cm} + \sum_{i=1}^{\log_2 \left(\frac{n-T^{*}}{\alpha}\right)}  4\exp{\left(- \alpha f^{2}_{1}(\alpha_1,1-\alpha_1)/2A^2\right)^{i}}
\nonumber \\ &
 \stackrel{(c)}{\leq} \frac{4\exp{\left(- \alpha \left(KL(P_1||P(1-\alpha_1))\right)^2/2A^2\right)}}{1-\exp{\left(- \alpha \left(KL(P_1||P(1-\alpha_1))\right)^2/2A^2\right)}}
   \nonumber \\ & \hspace{1cm} + \frac{4\exp\left( -\alpha f_1^{2}(\alpha_1,1-\alpha_1)/2A^2\right)}{1-\exp\left( -\alpha f_1^{2}(\alpha_1,1-\alpha_1)/2A^2\right)}
\label{eq: theorem-2-eqn-11}
\end{align}
From \eqref{eq: theorem-2-eqn-8} and \eqref{eq: theorem-2-eqn-10}, the following inequalities hold:
\begin{align}
  & \exp{\left(- \alpha \left(KL(P_1||P(1-\alpha_1))\right)^2/2A^2\right)} < \frac{1}{4} \nonumber \\ & \exp\left( -\alpha f_1^{2}(\alpha_1,1-\alpha_1)/2A^2\right) < \frac{1}{4}
\label{eq: theorem-2-eqn-12}
\end{align}
Thereby substituting in \eqref{eq: theorem-2-eqn-11} we get, 
\begin{align}
 P(\underset{t \in \mathcal{R}}{{\max}}\text{ } & S^{T_{\text{split}}}_{\text{DR}}(t) - S^{T_{\text{split}}}_{\text{DR}}(T^{*}) > 0) \nonumber \\ &  \leq \frac{16}{3}\left[\exp{\left(- \alpha \left(KL(P_1||P(1-\alpha_1))\right)^2/2A^2\right)} \right. \nonumber \\ & \hspace{1cm} + \left. \exp{\left(- \alpha f^{2}_{1}(\alpha_1,1-\alpha_1)/2A^2\right)}\right] \nonumber \\
& \leq \frac{32}{3}\exp{\left(- \alpha C ^2/2A^2\right)} 
\end{align}
where, $C = \min\{KL(P_1||P(1-\alpha_1)),f_{1}(\alpha_1,1-\alpha_1)\}$. 
 The analysis when $T_{\text{split}} \geq T^{*}$ follows in a similar manner, wherein we can show the following for the DR-CUSUM statistic $S^{T_{\text{split}}}_{\text{DR}}(t)$:
\begin{align}
P(\underset{t \in \mathcal{R}}{{\max}}\text{ } S^{T_{\text{split}}}_{\text{DR}}(t)-S^{T_{\text{split}}}_{\text{DR}}(T^{*}) > 0) & \leq \frac{32}{3}\exp{\left(- \alpha C ^2/2A^2\right)} 
\end{align}
where, $C = \min\{f_{2}(\alpha_2,\alpha_2),KL(P_2||P(\alpha_2)\}$. 
This completes the proof of Theorem \ref{thm: Theorem-2}.

\section{Overview on Density Ratio Estimators (DRE) \label{sec: Methods for Estimating Likelihood Ratios}}
In this section, we present a brief overview of data driven approaches for density ratio estimation (DRE) \cite{sugiyama2008direct}. The basic problem statement is the following: we are given samples from two distributions $P_{\text{left}}$ and $P_{\text{right}}$ and the goal is to estimate the density ratio functional $w(x) = {P_{\text{left}}(x)}/{P_{\text{right}}(x)}$. The main idea behind DRE is to use a parametric model for the density ratio functional, and then to learn the parameters using only samples from $P_{\text{left}}$ and $P_{\text{right}}$ in a principled manner. 

\noindent For instance, kernels, feed-forward or convolutional neural network (CNNs) can be used to model the density ratio estimator ($\hat{w}(x)$) \cite{sugiyama2008direct}\cite{kanamori2009least} \cite{nam2015direct}. Kernel based estimators are typically preferred low-dimensional data, while NNs have been shown to give better performance otherwise \cite{nam2015direct}.
We next describe one such principled approach \cite{sugiyama2008direct} (Kullback-Leibler Importance Estimation Procedure (KLIEP)) for DRE. Given the estimated density ratio $\hat{w}(x)$ (output of DRE model), if we know the true density value $P_{\text{right}}(x)$, we can estimate density $P_{\text{left}}(x)$ as follows,
\begin{align}
\label{eq: Estimated Density p_1(x)}
\hat{P}_{\text{left}}(x) = \hat{w}(x) {P}_{\text{right}}(x)
\end{align} 
For training the DRE model, KLIEP suggests minimizing the KL divergence between $P_{\text{left}}(.)$ and estimated density $\hat{P}_{\text{left}}(.)$, which simplifies to:
\begin{align}
\label{eq: KL Divergence True and Estimated Densities}
KL (P_{\text{left}} || \hat{P}_{\text{left}}) 
  \stackrel{\text{(i)}}{=}& \int_{x} P_{\text{left}}(x) \log \frac{P_{\text{left}}(x)}{\hat{w}(x)P_{\text{right}}(x)} dx \nonumber \\= & KL(P_{\text{left}}||P_{\text{right}})- \int_{x} P_{\text{left}}(x) \log \hat{w}(x) dx \nonumber\\= & KL(P_{\text{left}}||P_{\text{right}})- \mathbb{E}_{x\sim P_{\text{left}}}\left[\log (\hat{w}(x))\right] 
\end{align}
where, $\text{(i)}$ follows from definition of KL divergence and by using (\ref{eq: Estimated Density p_1(x)}). 
In (\ref{eq: KL Divergence True and Estimated Densities}), the first term is a constant w.r.t. DRE model parameters. Hence, the minimization of the KL-divergence in (\ref{eq: KL Divergence True and Estimated Densities}) is equivalent to maximization of $\mathbb{E}_{x \sim P_{\text{left}}} \left[ \log(\hat{w}(x))  \right]$.
Furthermore, for the estimated $\hat{P}_{\text{left}}$ to be a valid density, it must satisfy 
\begin{align}
\vspace{-5pt}
\label{eq: KL Divergence True and Estimated Densities Constraints}
\int_x \hat{P}_{\text{left}}(x) = \int_x \hat{w}(x) P_{\text{right}}(x)= \mathbb{E}_{x \sim P_{\text{right}}} \left[ \hat{w}(x)  \right] = 1
\end{align}
Using Lagrange parameter $\lambda$ to satisfy the constraints, we obtain the constrained optimization for the DRE as follows,
\newline \textbf{KLIEP Objective:}
\begin{align}
\label{eq: KL Divergence Neural Network Objective}
\max_{\hat{w}} \left(\mathbb{E}_{x \sim P_{\text{left}}} \left[\log(\hat{w}(x))\right] - \lambda \left(\mathbb{E}_{x \sim P_{\text{right}}} \left[ \hat{w}(x)  \right] - 1 \right)\right)&
\end{align}
By replacing the expectations by the sample means over $P_{\text{left}}$ and $P_{\text{right}}$, one can then use gradient ascent based optimization to find $\hat{w}$.  We want to highlight that there are several other approaches for density ratio estimation \cite{liu2017trimmed,yamada2011relative,kanamori2009least}. For instance, an unconstrained optimization (Least Squares Importance Fitting (LSIF)) is obtained in \cite{kanamori2009least} (also see \cite{nguyen2010estimating}) by minimizing the least square loss between the actual and estimated density ratios.
\newline \textbf{LSIF Objective:}
\begin{align}
\label{eq: LS Neural Network Objective}
\min_{\hat{w}} \left(\mathbb{E}_{x \sim P_{\text{left}}} \hat{w}(x)^2 - 2\mathbb{E}_{x \sim P_{\text{right}}}  \hat{w}(x) \right)&
\end{align}
Any model (neural network or kernels) for density ratio estimation can be leveraged in principle, for the design of our DRE-CUSUM change detection framework. 

 \noindent \underline{\textbf{Implementing DRE models}}
\newline In practice, we can only determine the empirical approximation to the objective functions in \eqref{eq: KL Divergence Neural Network Objective}, \eqref{eq: LS Neural Network Objective} (for both KLIEP and LSIF).
Given, samples $X_{[1:T_{\text{split}}]} \sim P_{\text{left}}$ and $X_{[T_{\text{split}}:n]} \sim P_{\text{right}}$. 
To estimate $P_{\text{left}}(.)/P_{\text{right}}(.)$, we sample $N_1$ samples from $X_{[1:T_{\text{split}}]}$ and $N_2$ samples from $X_{[T_{\text{split}}:n]}$ in each iteration, for training a density ratio estimator model denoted by $\hat{w}$.
For implementation, KLIEP objective in \eqref{eq: KL Divergence True and Estimated Densities Constraints} is empirically computed as follows: 
\begin{align}
\label{eq: emprical KLIEP}
\frac{1}{N_1} \sum_{x \sim P_{\text{left}}} \log(\hat{w}(x)) - \lambda \left( \frac{1}{N_2} \left(\sum_{x \sim P_{\text{right}}}  \hat{w}(x) \right)  - 1 \right)
\end{align}
Likewise, the LSIF objective in \eqref{eq: LS Neural Network Objective} can be empirically computed as:
\begin{align}
\label{eq: emprical LSIF}
\frac{1}{N_1} \sum_{x \sim P_{\text{left}}} \hat{w}(x)^2 - \frac{2}{N_2} \sum_{x \sim P_{\text{right}}}  \hat{w}(x)
\end{align}
\begin{algorithm}[!h]
	\caption{Training density ratio estimator (DRE) models} 
	\label{alg: Algorithm_DRE_train} 
	\begin{algorithmic}
		\STATE{ \textbf{INPUT:}  where $X_{[1:T_{\text{split}}]} \sim P_{\text{left}}$ and $X_{[T_{\text{split}}:n]} \sim P_{\text{right}}$}
		\STATE{ \textbf{OUTPUT:}  Estimate of $P_{\text{left}}(.)/P_{\text{right}}(.)$}
		\FOR{number of epochs}
		\STATE{1. Sample a mini-batch of $N_1$ samples from $X_{[1:T_{\text{split}}]}$ and $N_2$ samples from $X_{[T_{\text{split}}:n]}$.}
		\STATE{2. Compute objective in \eqref{eq: emprical KLIEP}, which is
        \begin{align*}
        \frac{1}{N_1} \sum_{x \sim P_{\text{left}}} \log(\hat{w}(x)) - \lambda \left( \frac{1}{N_2} \left(\sum_{x \sim P_{\text{right}}}  \hat{w}(x) \right)  - 1 \right)
        \end{align*}
        \STATE{3. Update parameters of the DRE model (for example, neural network weights) to maximize the above objective using gradient descent.}
		}
		\ENDFOR
	\end{algorithmic}
\end{algorithm}
\newline\noindent We summarize the training of DRE model using the KLIEP objective in \eqref{eq: emprical KLIEP} in Algorithm \ref{alg: Algorithm_DRE_train}, although one could use adapt Algorithm \ref{alg: Algorithm_DRE_train} to LSIF objective by optimizing the parameters of the DRE model to minimize \eqref{eq: emprical LSIF}.
  We next describe: the architecture of DRE models, the training of the DRE models, and details on synthetic dataset generation used for the results in the main paper.  

\subsection{DRE Architectures used in the paper}
For the scope of the experiments using DRE modeled using neural network, the neural network architectures are tabulated in Table \ref{table:architecture}.
For the kernel-based DRE, we use the package provided in \cite{pypi2016pypi}.

\begin{table*}[!t]
\centering 
\begin{tabular}{lll} 
\hline 
\textit{Experiment}& \textit{DRE model} & \textit{Architecture details}\\ [0.75ex] 
\hline 
Synthetic & Feedforward   & 4 dense layers \\
datasets &  neural network   & Hidden layer activation: Sigmoid \\
\,& DRE & Final layer activation: Softplus \\
\hline
Real-world & Kernel based   & Kernel type: Gaussian \\
datasets & DRE   & \cite{pypi2016pypi} \\
(USC, HASC) & \,  & \,\\
\hline 
Video   &  Convolutional  & 4 convolutional layers \\
datasets  & neural network & Hidden layer activation: Sigmoid\\
\, & DRE  & Final layer activation: Softplus \\
\hline 
\end{tabular}
\caption{Neural network DRE architecture details.} 
\label{table:architecture} 
\end{table*}

 \noindent \textit{Details of experiment to test robustness of DRE-CUSUM:} In the Fig. 5b (in the paper) we have the $10-$dimensional time-series data  $X_{[1:1000]}$ with change point $T^{*} = 350$, such that $X_{[1:350]} \sim P_1$ and $X_{[350:1000]} \sim P_2$, where $P_1 \sim \mathcal{N}(\vec{\mu}_1,I)$ and $P_2 \sim \mathcal{N}(\vec{\mu}_2,I)$.
Entries of mean-vector $\vec{\mu}_{1}$ is sampled from $\text{Unif.}[-1,1]$, while $\vec{\mu}_{2}$ is varied by adding (small) increments to $\vec{\mu}_{1}$.
We set $T_{\text{split}} = 500$.
Consequently, for this example, we have $P_{\text{left}} = \alpha^{T_{\text{split}}} P_1 + (1-\alpha^{T_{\text{split}}})P_2$, where $\alpha^{T_{\text{split}}} = 0.7$ (i.e. $=350/500$).
However, we have $P_{\text{right}} = P_2$. 
In each epoch, we sample a minibatch of data from both $P_{\text{left}}$ and $P_{\text{right}}$, and train the DRE model with either using \eqref{eq: emprical KLIEP} or \eqref{eq: emprical LSIF} as the training objective.
Per table \ref{table:architecture}, we use a feed-forward neural network with $10$- input nodes. The width of the three hidden layers are $256$, $512$, $128$, respectively, and the output of the neural network DRE is the estimated density ratio $\hat{w}(x)$ corresponding to the input sample $x$.
During training, we set set the size of the mini-batch to be 64, and train the neural network DRE for 500 iterations.
We train the neural network DRE using KLIEP objective in \eqref{eq: emprical KLIEP}. 
Post-training, we compute the density ratio $\hat{w}(x)$, $\forall x \in X_{[1:1000]}$, and subsequently plot the DRE-CUSUM statistic which is depicted in Fig. 5b in the paper.  
\noindent \textit{We refer the readers to \href{https://www.dropbox.com/sh/xkchf2iajge57jq/AACztXuP-W16VSkUuoTcNY0ya?dl=0}{{\color{magenta}this link}} (which was provided in the paper) which contains the code to generate the results}. 

\noindent \textit{Details of experiment in Table 1 of the paper:}
We generate a 50-dimensional Gaussian time-series data of length $2000$ with change points at time-instances $t = \{150,200,450,525,700,725,1200\}$.
We set the co-variance matrix across all segments in the time-series data to be the same, and is generated as follows: each entry in standard deviation vector $\sigma$ is sampled from a distribution $\text{Unif.}[1,3]$. 
We obtain covariance matrix $\Sigma = \sigma^{T}\sigma I$.
However, we vary the mean vector across the different segments in the time-series data, such that the entries of the mean vector across different segments (ordered) are sampled from: (i) $\text{Unif.}[-1,1]$, (ii) $\text{Unif.}[-2,2]$, (iii) $\text{Unif.}[-3,3]$,  (iv) $\text{Unif.}[-4,4]$, (v) $\text{Unif.}[-3,3]$, (vi) $\text{Unif.}[-10,10]$, (vii) $\text{Unif.}[-20,20]$, (viii) $\text{Unif.}[-1,1]$.
The architecture used in this experiment is a feed-forward neural network with $50$ input nodes. 
The hidden layer widths (from input layer towards output) are $256$, $512$, $128$, and the neural network based DRE is trained for $500$ iterations. 
The experiment is performed using both KLIEP and LSIF objectives.

\section{Additional experimental results on video datasets}
\label{sec: experiments-appendix}
\begin{figure}[!h]
\centering
\begin{subfigure}[b]{0.46\textwidth}
   \includegraphics[width=1\linewidth]{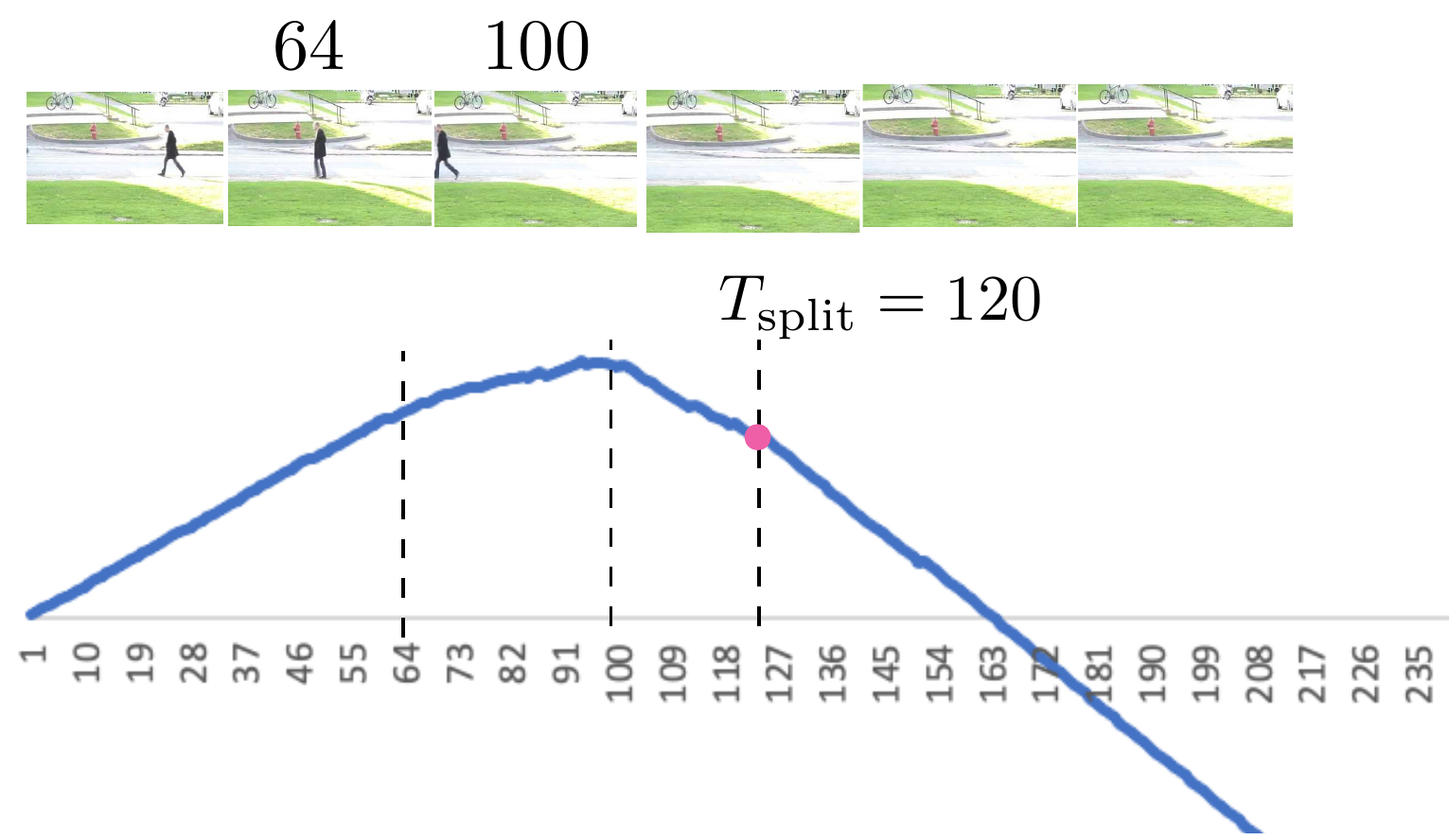}
\caption{DRE-CUSUM on Pedestrian dataset. \label{fig:  DRE-CUSUM-Pedestrian} }
\end{subfigure}

\begin{subfigure}[b]{0.46\textwidth}
   \includegraphics[width=1\linewidth]{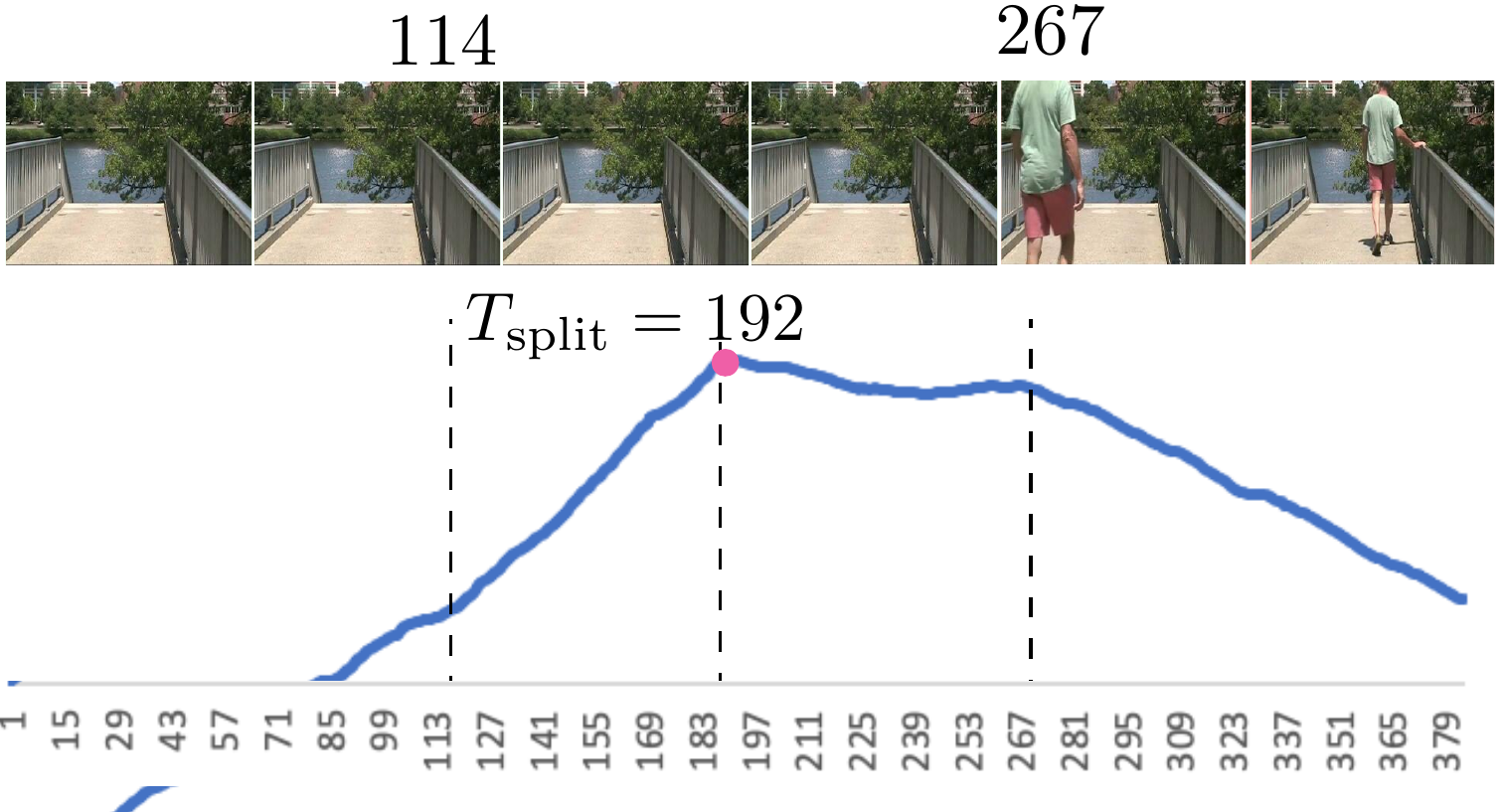}
   \caption{DRE-CUSUM algorithm on Overpass dataset. \label{fig:  DRE-CUSUM-Overpass-2}} 
\end{subfigure}
\caption{Video event detection using DRE-CUSUM for detecting entry and exit instances of a person.}
\vspace{-4pt}
\end{figure}

We conducted additional experiments using DRE-CUSUM on real-world video data in 2012-Dataset \cite{goyette2012changedetection}. 
In particular, the objective was to perform activity detection (in particular, detect the entry/exit of a person) in the sequence of video frames.
We present the results on pedestrian and overpass video sequences present in  the 2012-Dataset \cite{goyette2012changedetection}.  
In this experiment, with a time-series of $240$ frames, a person is present in frames $0-100$. As shown in Fig. \ref{fig:  DRE-CUSUM-Pedestrian}, we first set $T_{\text{split}} = 120$.
We observe slope changes at frames $65$ and $100$.
It can be noted that, the video frames $65-100$ belong to the transition period when the person gradually exists and is no more present in the video.
As we can observe, the DRE-CUSUM statistic is able to detect both the beginning and end of these transition frames. 

 In the second experiment, we consider a time-series of $385$ frames. The person appears in the $260^{th}$ frame. We first set $T_{\text{split}}=192$ and obtain the corresponding DRE-CUSUM statistic as shown in Fig. \ref{fig:  DRE-CUSUM-Overpass-2}.  Slope changes at are observed at instances corresponding to frames $192$ (i.e. $T_{\text{split}}$), and $267$.
 The slope change at around frame $120$ corresponds to a false alarm (upon visual inspection no change is observed).

 \textit{Additional Architectural details:} In general for event detection experiments, the architecture in Table \ref{table:architecture} is suitable.
 In the hidden layers of the convolutional neural network based DRE, we apply max-pooling, and the KLIEP objective is used to train the parameters of the neural network.   
 We train the neural network DRE for $2000$ iterations. 

\end{document}